\documentclass{article} 

\PassOptionsToPackage{table}{xcolor}
\usepackage{iclr2027_conference,times}
\usepackage{cmlab_arxiv} 

\usepackage[utf8]{inputenc} 
\usepackage[T1]{fontenc}    
\usepackage{hyperref}       
\usepackage{etoc}
\usepackage{url}            
\usepackage{booktabs}       
\usepackage{amsfonts}       
\usepackage{nicefrac}       
\usepackage{microtype}      
\usepackage{xcolor}         
\usepackage{graphicx}
\usepackage{float}
\usepackage{placeins}
\usepackage{amsmath, amssymb}
\usepackage{dsfont}
\usepackage{algorithm}
\usepackage{algpseudocode}
\usepackage{makecell}
\usepackage{wrapfig}
\usepackage{bbm}
\definecolor{refnavy}{HTML}{0D207F}

\hypersetup{
  colorlinks=true,
  linkcolor=refnavy,
  citecolor=refnavy,
  urlcolor=refnavy,
  pdfborder={0 0 0},
}

\usepackage[capitalize,noabbrev]{cleveref}

\crefname{figure}{Fig.}{Figs.}
\Crefname{figure}{Figure}{Figures}
\crefname{table}{Tab.}{Tabs.}
\Crefname{table}{Table}{Tables}
\crefname{section}{Sec.}{Secs.}
\Crefname{section}{Section}{Sections}
\crefname{subsection}{Sec.}{Secs.}
\Crefname{subsection}{Section}{Sections}
\crefname{subsubsection}{Sec.}{Secs.}
\Crefname{subsubsection}{Section}{Sections}
\crefname{equation}{Eq.}{Eqs.}
\Crefname{equation}{Equation}{Equations}
\crefname{algorithm}{Algo.}{Algos.}
\Crefname{algorithm}{Algorithm}{Algorithms}

\definecolor{cmlabPrefix}{HTML}{0D207F}
\definecolor{cmlabNumber}{HTML}{0D207F}
\newcommand{\cmlabRef}[4]{%
  {\color{cmlabPrefix}#1}~#3{\color{cmlabNumber}#2}#4%
}
\newcommand{\cmlabRefParen}[4]{%
  {\color{cmlabPrefix}#1}~#3{\color{cmlabNumber}(#2)}#4%
}
\crefformat{figure}    {\cmlabRef{Fig.}{#1}{#2}{#3}}
\Crefformat{figure}    {\cmlabRef{Figure}{#1}{#2}{#3}}
\crefformat{table}     {\cmlabRef{Tab.}{#1}{#2}{#3}}
\Crefformat{table}     {\cmlabRef{Table}{#1}{#2}{#3}}
\crefformat{section}   {\cmlabRef{Sec.}{#1}{#2}{#3}}
\Crefformat{section}   {\cmlabRef{Section}{#1}{#2}{#3}}
\crefformat{subsection}{\cmlabRef{Sec.}{#1}{#2}{#3}}
\Crefformat{subsection}{\cmlabRef{Section}{#1}{#2}{#3}}
\crefformat{equation}  {\cmlabRefParen{Eq.}{#1}{#2}{#3}}
\Crefformat{equation}  {\cmlabRefParen{Equation}{#1}{#2}{#3}}

\usepackage{graphicx}         
\usepackage{wrapfig,lipsum}

\usepackage{booktabs}
\usepackage{makecell}
\usepackage[table]{xcolor}
\usepackage{amsmath}
\usepackage{multirow}
\usepackage{xcolor}
\usepackage{arydshln}
\usepackage{titletoc}

\definecolor{Pose}{HTML}{4DB6AC}    
\definecolor{Domain}{HTML}{FFB74D}  
\definecolor{Dist}{HTML}{BA68C8}    

\usepackage{makecell}

\usepackage{pifont}                

\usepackage{nicematrix}

\definecolor{purple}{RGB}{150, 0, 210} 

\definecolor{myred}{RGB}{132, 10, 50} 

\definecolor{my}{RGB}{220, 53, 98} 

\definecolor{dh}{RGB}{250, 13, 10}

\definecolor{bluejh}{RGB}{0, 0, 0}

\definecolor{greenjh}{RGB}{0, 0, 0}

\definecolor{royalbluejh}{RGB}{0, 0, 0}

\definecolor{royalbluejhxx}{RGB}{0, 0, 0}

\definecolor{royalbluejhxxx}{RGB}{0, 0, 0}

\definecolor{royalbluejhx}{RGB}{0, 0, 0}

\definecolor{royalbluejhy}{RGB}{0, 0, 200}

\makeatletter
\newcommand{\sidecaptiontable}[2]{%
  \begingroup
  \def\@captype{table}%
  \caption{#1}%
  \label{#2}%
  \endgroup
}

\newcommand{\sidecaptionfigure}[2]{%
  \begingroup
  \def\@captype{figure}%
  \caption{#1}%
  \label{#2}%
  \endgroup
}
\makeatother 

\title{Chameleon: Style-Content Disentangled Framework for Cross-Domain Object Compositing}

\author{Sukhun Ko, Jeahun Sung, Soo Ye Kim, Jihyong Oh}

\cmlabAuthors{Sukhun Ko$^{1}$ \qquad Jeahun Sung$^{1}$ \qquad Soo Ye Kim$^{\dagger,2}$ \qquad Jihyong Oh$^{\dagger,1}$}
\cmlabAffiliations{$^{1}$ CMLab, Chung-Ang University \qquad $^{2}$ Adobe Research}
\cmlabAuthorEmail{\{looloo330,jhseong,jihyongoh\}@cau.ac.kr \qquad sooyek@adobe.com}
\cmlabProjectPage{https://cmlab-korea.github.io/Chameleon/}
\begin{document}

\maketitle

{
  \renewcommand{\thefootnote}{\fnsymbol{footnote}}
  \footnotetext[1]{Co-Corresponding authors.}
}

\vspace{-30pt}
\begin{figure}[H]
    \centering
    \includegraphics[width=\linewidth]{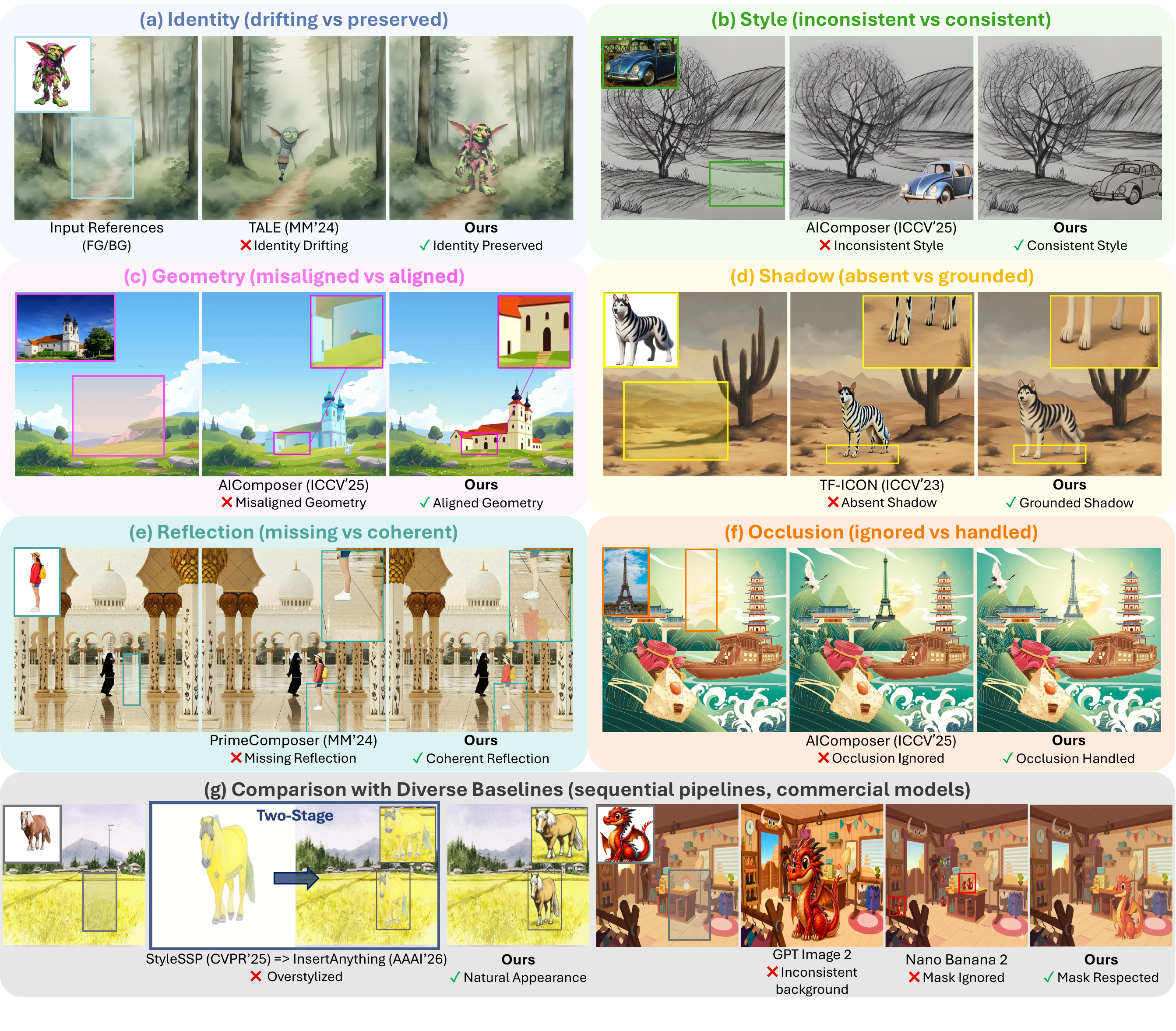}
    \vspace{-10mm}
    \caption{\textbf{Cross-domain compositing results by our Chameleon:} (a) shows that Chameleon preserves foreground identity, while (b) maintains consistent style. (c) and (d) demonstrate compositional plausibility via aligned geometry and grounded shadows, respectively. (e) demonstrates scene-aware compositing via coherent reflections, while (f) handles foreground-background occlusion relationships. (g) shows that Chameleon outperforms two-stage cascaded pipelines that combine style transfer and object insertion, which tend to over-stylize the foreground, as well as commercial models (GPT-Image-2~\citep{openai2026gptimage2}, Nano-Banana-2~\citep{google2026gemini3flashimage}) that shift background tone or ignore the input mask.} 
    \vspace{-10pt}
    \label{fig:fig1}
\end{figure}


\clearpage
\begin{abstract}
Image compositing aims to seamlessly insert a foreground object into a background image, and recent advances in diffusion models have significantly enhanced the quality, especially when the foreground and background images come from the \textit{same domain} (e.g., natural images). However, \textit{cross-domain} compositing, where the foreground and background come from different domains, is relatively underexplored and remains challenging because the model must preserve the foreground object's identity while stylizing it to match the background domain. Existing cross-domain compositing approaches largely rely on training-free blending and refinement strategies. This is partly due to the lack of large-scale paired datasets for cross-domain compositing, limiting the development of training-based solutions. As a result, they are limited to tone-level alignment and often produce style-inconsistent or overstylized results. To overcome such limitations, we construct ChameleonDataset, the first large-scale training dataset for cross-domain compositing, with a comprehensive evaluation benchmark, built through a scalable data construction pipeline. Building on this, we propose Chameleon, a novel two-stage training-based cross-domain compositing framework. In the first stage, we propose Signed Style-Content Graph Learning (SSCGL) to train ChameleonEncoder, which effectively disentangles style and content representations. In the second stage, we introduce Style Gap-Aware Anisotropic Modulation (SGAM) into a diffusion transformer for effective cross-domain compositing, adaptively regulating how style tokens from the first-stage ChameleonEncoder are injected according to the foreground-background style gap, spatial region, and diffusion timestep.
Our method outperforms state-of-the-art in-domain and cross-domain compositing models, sequential pipelines and commercial models,
achieving improvements in both compositional plausibility and stylistic fidelity.
\end{abstract}
\vspace{-5mm}
\etocdepthtag.toc{mtchapter}
\section{Introduction}
\label{sec:intro}
\vspace{-2mm}
Image compositing~\citep{song2023objectstitch,canet2024thinking} aims to place a reference object(s) in a foreground image (i.e., reference image) at a desired location within a background image (i.e., target image) such that the inserted object appears natural within the background canvas. With recent advances in diffusion models~\citep{ho2020denoising,song2021scorebased}, image compositing (i.e., generative object compositing) has significantly improved by re-synthesizing the composite region through learned generative priors~\citep{chen2024anydoor, huang2025dreamfuse}, rather than merely adjusting pixel values near the boundary~\citep{perez2023poisson}. This enables more natural integration in terms of appearance, geometry, and semantics, allowing the foreground to better align with the background.

These advances have made image compositing practical for real-world applications such as digital content creation and advertising, where the foreground and background typically come from the same photorealistic visual domain. However, another line of use cases is in creative ideation, where a user may wish to insert an asset coming from a different domain. For instance, a user may have a specific car photograph in mind that they imagine inserting into a rough first sketch, as shown in Fig.~\ref{fig:fig1} (b). Rather than needing to manually draw the car in a similar style, cross-domain compositing can help directly transform this car into a similar sketch style while inserting it to the desired background, making the ideation process more efficient and creative by expanding the pool of object and background assets to \textit{any} stylistic domain.

Nevertheless, cross-domain compositing is a challenging research problem that involves multiple requirements, as shown in Fig.~\ref{fig:fig1}: (a) preserving the identity of the foreground reference, (b) adapting its appearance to the style of the background, (c) maintaining geometric and contextual plausibility within the scene, and (d)-(f) handling shadows, reflections, and occlusions for visual coherence.
 
Among prior work in cross-domain compositing, early methods such as TF-ICON~\citep{lu2023tf} apply DDIM inversion~\citep{mokady2023null} and integrate foreground and background via latent-space blending and attention injection. TALE~\citep{pham2024tale} leverages intermediate latents instead of pure noise, improving structural and stylistic preservation. More recent methods such as AIComposer~\citep{li2025aicomposer} instead perform pixel-space blending and leverage CLIP features~\citep{ye2023ip} for foreground–background fusion.

Despite these attempts, existing methods fall short of fully solving this challenging problem, as they share a common paradigm that freezes a pre-trained text-to-image (T2I) diffusion model~\citep{rombach2022high} and performs compositing solely through blending-style refinement, leading to several limitations. First, these methods depend on user-provided prompts to drive a frozen T2I diffusion backbone, yet such prompts are difficult to craft and cannot fully match the foreground, leading to \textit{identity drift} (Fig.~\ref{fig:fig1} (a)), where the prompt ``a goblin'' fails to capture the specific identity of the foreground. Second, these blending-based methods apply statistical matching such as AdaIN~\citep{huang2017arbitrary} to align the foreground with the background distribution, yet this approach is limited to tone-level alignment and cannot bridge large domain gaps, resulting in \textit{style inconsistency} (Fig.~\ref{fig:fig1} (b)). Third, these methods lack training for cross-domain compositing, resulting in \textit{geometry misalignment} and \textit{absence of shadows} (Fig.~\ref{fig:fig1} (c,d)). These limitations stem from the lack of suitable training data for cross-domain compositing.
\begin{table}[t]
\centering
\vspace{-2mm}
\caption{
Comparison with existing stylization and compositing datasets. Ours is the only dataset supporting both tasks and view augmentation with real-image supervision (200K samples). 
}
\label{tab:dataset}

\footnotesize
\setlength{\tabcolsep}{2.5pt}
\renewcommand{\arraystretch}{0.85}

\begin{tabular}{l c c c c c}
\toprule
\textbf{Dataset}
& \textbf{Stylization}
& \textbf{Compositing}
& \textbf{View Augmentation}
& \textbf{Supervision}
& \textbf{\#Samples} \\
\midrule

OmniConsistency
& $\checkmark$
& $\times$
& $\times$
& {Synthetic}
& 2.6K \\

OmniStyle
& $\checkmark$
& $\times$
& $\times$
& {Synthetic}
& 150K \\

DreamFuse
& $\times$
& $\checkmark$
& $\times$
& {Synthetic}
& 84K \\

AnyInsertion
& $\times$
& $\checkmark$
& $\triangle$
& {Real}
& 159K \\

\textbf{Ours}
& \textbf{$\checkmark$}
& \textbf{$\checkmark$}
& \textbf{$\checkmark$}
& \textbf{Real}
& \textbf{200K} \\

\bottomrule
\end{tabular}
\vspace{-6mm}
\end{table}

To address these limitations, we construct ChameleonDataset$_{\text{tr}}$, the first large-scale training set for cross-domain compositing. Unlike prior datasets that rely on synthetic-image supervision (Tab.~\ref{tab:dataset}), we adopt a reverse pipeline driven by real-image supervision. This mitigates artifacts in stylized generations induced in the prior datasets (Sec.~\ref{sec:dist_shift}), and exposes the model to diverse compositional scenarios, enabling natural geometry alignment and shadow grounding. We further introduce ChameleonDataset$_{\text{ev}}$ that evaluates both compositional plausibility and stylistic fidelity under diverse real-world scenarios.

With training data in place, effective cross-domain compositing further requires disentangled content and style representations to preserve foreground identity while adapting to the background style. Inspired by signed graph learning~\citep{derr2018signed,huang2021sdgnn}, we propose ChameleonEncoder, a style-content disentangled encoder trained with our novel Signed Style-Content Graph Learning (SSCGL). Unlike prior signed graph representation learning methods~\citep{derr2018signed,huang2021sdgnn}, SSCGL leverages edge signs for factor disentanglement by assigning opposite signs to the same image pair across the style and content representation spaces. By structuring DINOv3~\citep{simeoni2025dinov3} features with these signed relations, SSCGL yields disentangled style and content tokens.
These tokens are then injected into the diffusion transformer (DiT), while our novel Style Gap-Aware Anisotropic Modulation (SGAM) adaptively regulates style injection according to the foreground-background style gap, spatial region, and diffusion timestep for effective cross-domain compositing.
In summary, our contributions are as follows:
\vspace{-2mm}
\begin{itemize}
 \setlength\itemsep{0.05cm}
  \item We construct ChameleonDataset$_{\text{tr}}$, the first large-scale dataset for cross-domain image compositing, built via a reverse data generation pipeline that can be applied on any images. It provides real-image supervision rather than synthetic-image supervision, mitigating generative artifacts and enabling faithful stylization.

  \item We introduce ChameleonDataset$_{\text{ev}}$, a comprehensive benchmark evaluating compositional plausibility (grounding, reflection) and stylistic fidelity (pixel art, stylized-to-stylized).
  
  \item We propose Chameleon, a novel two-stage training-based cross-domain compositing framework consisting of (i) ChameleonEncoder, a style-content disentangled encoder trained with Signed Style-Content Graph Learning (SSCGL), which structures DINOv3 features with signed graph relations to achieve the disentanglement essential for identity preservation and stylization, and (ii) Style Gap-Aware Anisotropic Modulation (SGAM), which adaptively regulates style injection into a DiT for effective cross-domain compositing.
\end{itemize}

\FloatBarrier
\vspace{-3mm}
\section{Related Work}
\label{sec:relatedwork}
\vspace{-3mm}
\noindent\textbf{Image compositing.} Naturally inserting a foreground image into a background image has long been a challenging problem in image editing~\citep{brooks2023instructpix2pix,meng2022sdedit}. Classical techniques~\citep{perez2023poisson, porter1984compositing} improved visual quality but remain limited to pixel-level operations, lacking semantic awareness. Recently, diffusion models have significantly improved object insertion by leveraging priors from large-scale data. For example, Paint-by-Example~\citep{yang2023paint} fills masked regions using a reference image but fails to preserve object identity, which AnyDoor~\citep{chen2024anydoor} mitigates via an ID extractor and a high-frequency map.

However, such methods struggle when the reference and background images belong to heterogeneous domains. To overcome this limitation, methods such as TF-ICON~\citep{lu2023tf}, TALE~\citep{pham2024tale}, and AIComposer~\citep{li2025aicomposer} have been proposed for cross-domain compositing, performing latent- or pixel-level blending based on a frozen text-to-image backbone~\citep{rombach2022high}. Despite improved cross-domain harmonization, these approaches still exhibit several limitations. For instance, without explicit supervision, it remains unclear how much blending should be applied, as excessive blending may degrade foreground identity, while insufficient blending may prevent the foreground from fully adapting to the background domain (Fig.~\ref{fig:fig1} (b)). Moreover, these methods rely on blending-based refinement rather than learning the insertion process, and thus struggle to jointly synthesize the inserted foreground and compositional effects such as shadows and reflections (Fig.~\ref{fig:fig1} (d),(e)).

\noindent \textbf{Style transfer.}
Style transfer has been widely studied and is relevant to cross-domain compositing in how visual style is represented and transferred across domains.
Early neural style transfer methods~\citep{gatys2016image} represent style through Gram statistics of pre-trained VGG features~\citep{simonyan2014very}, primarily capturing low-level appearance statistics such as color and texture. Recent diffusion-based approaches increasingly leverage semantic image encoders such as CLIP~\citep{radford2021learning,ye2023ip} for higher-level style conditioning.
More recent works have explored self-supervised features such as DINO~\citep{caron2021emerging,oquab2024dinov}, but primarily use them as content or semantic references, even in style-oriented applications~\citep{jia2026styledecoupler,zhou2024deformable}. In contrast, our analysis shows that DINOv3~\citep{simeoni2025dinov3} already contains substantial style information, while its style and content representations remain entangled (Tab.~\ref{tab:why_dinov3}). Such entanglement is particularly detrimental to cross-domain compositing, as background content can leak into the insertion region and interfere with foreground insertion (Fig.~\ref{fig:fig4}). We therefore fully leverage DINOv3 by disentangling its features into style and content representations via our Signed Style-Content Graph Learning (SSCGL), enabling separate style and content conditioning for cross-domain compositing.

\vspace{-3mm}
\section{Method}
\label{sec:method}  
\vspace{-3mm}
\subsection{ChameleonDataset}
\vspace{-5mm}
\label{sec:dataset}
\begin{figure}[H]
    \centering
    \includegraphics[width=\linewidth]{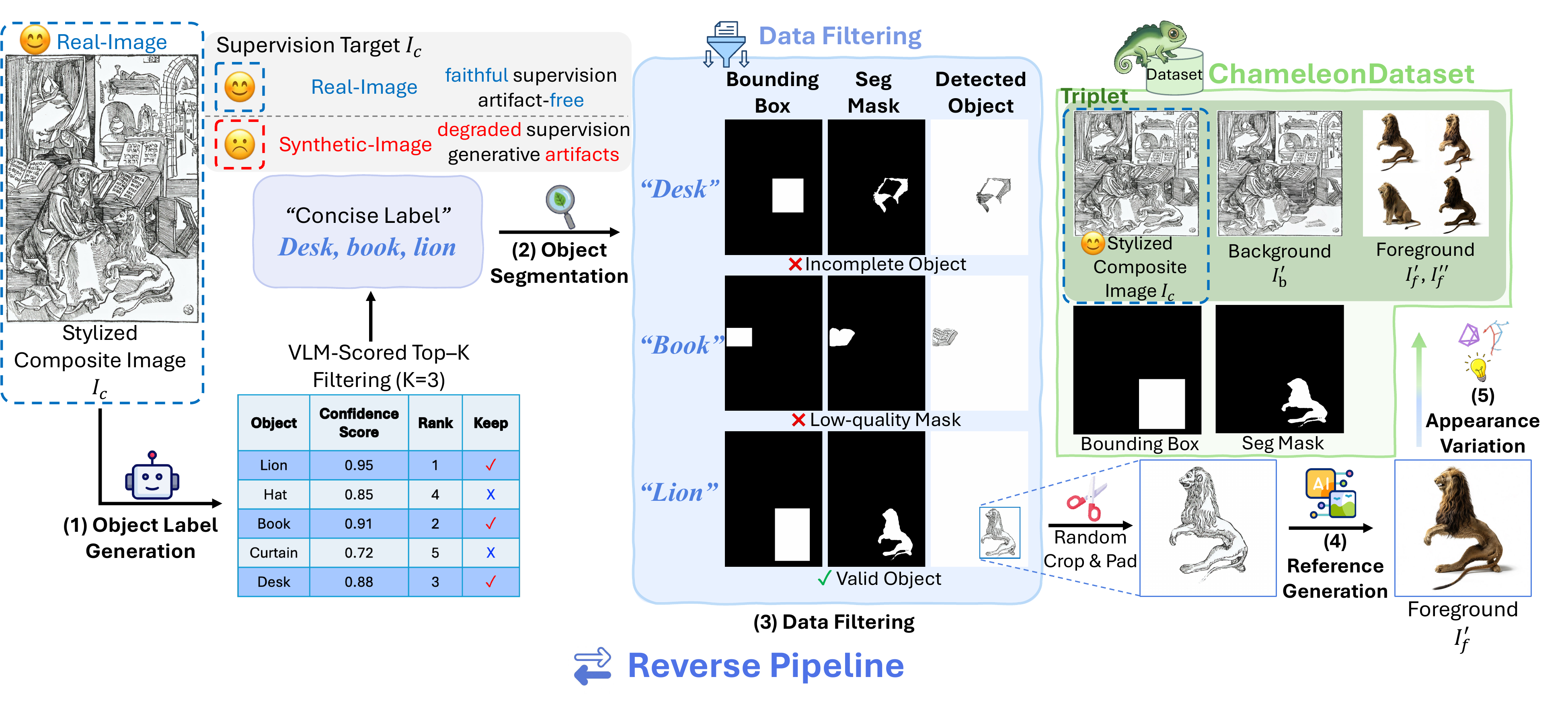}
    \vspace{-25pt}
    \caption{
    \textbf{Our reverse data generation pipeline.} Our reverse data generation pipeline starts from real stylized composite images $I_c$, ensuring \textit{faithful, artifact-free supervision}.
    }
    \vspace{-20pt}
    \label{fig:fig2}
\end{figure}
Cross-domain compositing can be formulated over triplets $(I_f, I_b, I_c)$, where a foreground image $I_f$ is placed at a target location in a background image $I_b$ and harmonized to the style of $I_b$, yielding the stylized composite $I_c$. Supervision is applied to $I_c$, so whether $I_c$ follows a synthetic- or real-image distribution becomes critical {(Sec.~\ref{sec:dist_shift})}. Throughout the paper, $I_c$ denotes a real-image composite, and $I_c'$ or $I_c''$ its synthetic counterpart, where $'$ denotes the synthetic output of a generative model. 


\textbf{Forward pipeline.} Prior approaches~\citep{huang2025dreamfuse, yang2023paint, canberk2024erasedraw} adopt a forward data generation pipeline that constructs the supervision target $I_c'$ or $I_c''$ from synthetic generations (Fig.~\ref{fig:fig10} in \textit{Appendix}). Using such degraded supervisions leads to the model learning to reproduce that synthetic distribution rather than the real-image one, yielding an inherently suboptimal mapping regardless of model capacity.

\noindent\textbf{Reverse pipeline.}
We address these limitations by inverting the construction process. Rather than \textit{generating} stylized composites, our \emph{reverse pipeline} (Fig.~\ref{fig:fig2}) starts from stylized real-image composites $I_c$ drawn from curated stylization datasets~\citep{li2024styletokenizer, liao2022artbench, ju2023human} and treats them as ground-truth $I_c$. Given each $I_c$, we segment its salient foreground objects and, for each, synthesize a foreground image $I_f'$ and remove the corresponding object from $I_c$ using an off-the-shelf object removal model~\citep{zhao2026precise} to obtain $I_b'$, yielding one training triplet $\{I_f', I_b', I_c\}$ per object. Here, $I_c$ is a real image preserved from the source, while $I_f'$ and the removed region of $I_b'$ contain synthetic content and are used only as input conditions. When constructing $I_b'$, we retain the original pixels of $I_c$ outside the segmentation mask to preserve the intact regions of the original background, such that only the removed foreground region is synthesized. Crucially, generative artifacts in $I_f'$ and the removed region of $I_b'$ do not affect the output distribution of the learned model, since supervision is applied only to $I_c$, which remains free of generative artifacts by construction. Our pipeline consists of five stages (Fig.~\ref{fig:fig2}).
\textbf{(1) Object label generation.} We first employ Qwen3-VL~\citep{bai2025qwen3} as a concise noun generator that produces object labels compatible with SAM3~\citep{carion2026sam}. 
\textbf{(2) Object segmentation.} Using the filtered labels, SAM3 produces foreground candidates from $I_c$, each represented as a segmented region with a corresponding mask.
\textbf{(3) Data filtering.} We then prompt Qwen3-VL~\citep{bai2025qwen3} again with a filtering instruction to score each candidate along multiple criteria and produce a binary keep/reject decision (see \textit{Appendix} for details).
\textbf{(4) Reference generation.} Each valid candidate is passed through a reference generation model~\citep{wu2025qwen} to obtain $I_f'$. Although $I_f'$ is itself a synthetic-image, it is only used as input. Supervision thus remains \textit{faithful} and \textit{artifact-free} on the real-image $I_c$.
\textbf{(5) Appearance variation.} Finally, we apply appearance variation generation~\citep{wu2025qwen} to a subset of the resulting $I_f'$, perturbing camera parameters (e.g., azimuth, elevation) to obtain $I_f''$ with diverse poses and lighting. This forces the model to match $I_f''$ with the pose and lighting of $I_c$ rather than naively pasting it. Construction details and stage-wise statistics are provided in Appendix~\ref{sec:dataset_statistics}.
\begin{figure}[H]
    \centering
    \includegraphics[width=\linewidth]{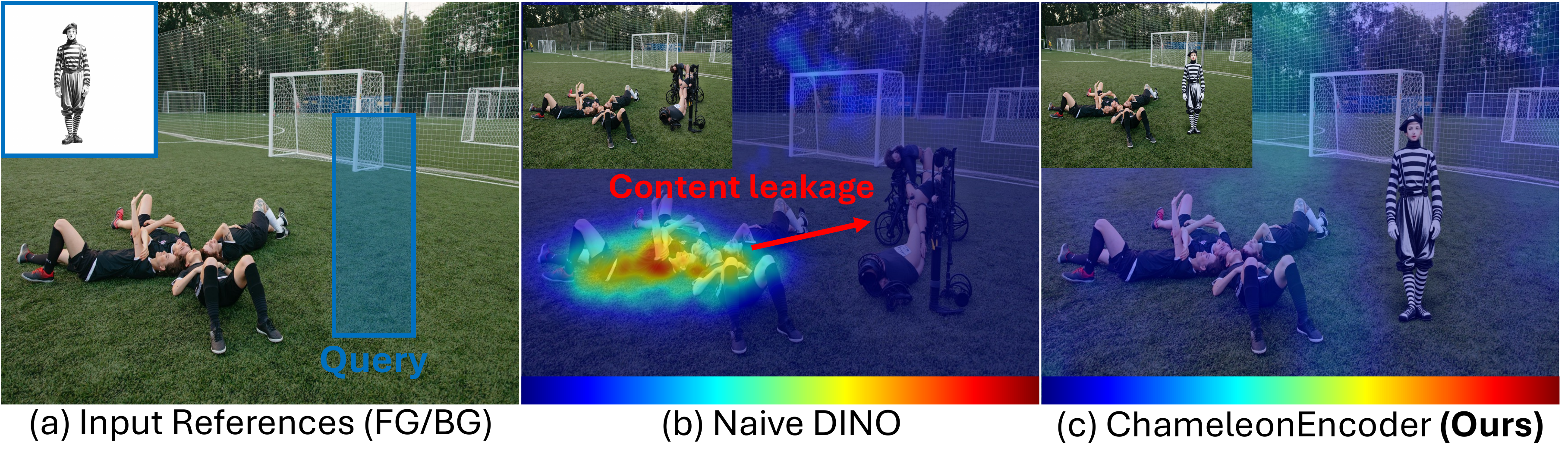}
    \vspace{-5mm}
    \caption{
    We visualize attention from the query region (a) to the background, where red indicates higher attention. (b) Naive DINO focuses on human regions, causing \textit{content leakage}. (c) ChameleonEncoder (\textit{style} head) attends globally, capturing background \textit{style} rather than \textit{content}, while inserting the foreground.}
    \label{fig:fig4}
\end{figure}
\vspace{-3mm}\vspace{0cm}
\label{sec:stage1}
\subsection{Design Considerations for Chameleon}
\vspace{-3mm}
\label{sec:method_motivation}
The goal of cross-domain compositing is to composite $I_f$ onto $I_b$ across heterogeneous domains while preserving the identity of $I_f$ and adapting its style to $I_b$. Meeting these requirements raises two key design considerations: \emph{role-specific representations} and \emph{adaptive style injection}. Together, these considerations motivate a two-stage framework with complementary roles: Stage~1 learns role-specific representations, and Stage~2 uses them to adaptively inject the resulting style representation.

\noindent\textbf{Role-specific representations.}
Preserving foreground identity while adapting it to the background style requires disentangled, role-specific representations: \textit{content} from $I_f$ and \textit{style} from $I_b$. Yet off-the-shelf encoders~\citep{radford2021learning,tschannen2025siglip} entangle the two, potentially leaking foreground style or background content into the composite (Fig.~\ref{fig:fig4}). To obtain representations specialized for these distinct roles, \textbf{Stage~1} (Sec.~\ref{sec:stage1}) pre-trains \emph{ChameleonEncoder}, a style-content disentangled encoder on a DINOv3 backbone via Signed Style-Content Graph Learning (SSCGL). 

\noindent\textbf{Adaptive style injection.}
Given these disentangled representations, the next challenge is to properly inject the style signal into the DiT. The central difficulty lies in calibrating, through the DINO style tokens, \emph{how much}, \emph{where}, and \emph{when} to inject style. Accordingly, \textbf{Stage~2} (Sec.~\ref{sec:stage_generation}) uses ChameleonEncoder's disentangled representations to condition the DiT, while reusing its Stage~1 style head to compute the foreground--background style gap from $I_f$ and $I_b$. We introduce Style Gap-Aware Anisotropic Modulation (SGAM), which adaptively modulates style attention according to this gap, spatial region, and diffusion timestep, without disturbing other tokens.
\subsection{Stage 1 -- ChameleonEncoder: Style-Content Disentangled Encoder}
\noindent\textbf{Leveraging Semantic Encoders for Style and Content Embeddings.}\par
Cross-domain compositing requires an encoder that provides role-specific signals: foreground content to preserve object identity and background style to adapt the inserted foreground to match the appearance of $I_b$. Recent self-supervised visual encoders such as DINOv3~\citep{simeoni2025dinov3} have demonstrated strong transferability and have been increasingly adopted across diverse downstream tasks, including fine-grained retrieval~\citep{liu2026efficient}, open-vocabulary detection~\citep{Sun_2026_CVPR} and 3D visual understanding~\citep{ehm2026teaching}, making them a compelling source of \textit{semantic} features. However, these applications primarily exploit DINO features for semantic or structural information. Even in style-oriented applications, DINO features are primarily used as content or semantic references~\citep{jia2026styledecoupler,zhou2024deformable}.
We therefore examine whether DINOv3 can provide not only semantic information but also style-relevant information required for cross-domain compositing. Our analysis in Tab.~\ref{tab:why_dinov3} shows that DINOv3 captures stronger style-related similarity than other off-the-shelf semantic encoders, yet its style and content representations remain entangled. To probe this, we feed $I_f$ (Fig.~\ref{fig:fig4} (a) top-left) and $I_b$ to the DiT as VAE latents, and condition on (raw) DINOv3 tokens extracted from $I_b$. As shown in Fig.~\ref{fig:fig4}, the model generates background people in the insertion region instead of the foreground object due to the content information retained in the raw DINOv3 embeddings. We term this failure \emph{content leakage}, where background content is transferred together with style. Taken together, these observations indicate that DINOv3 contains substantial style-relevant information, but this information remains entangled with content, motivating explicit style-content disentanglement.

\begin{figure}[t]
    \centering
    \vspace{-4mm}
    \includegraphics[width=\linewidth]{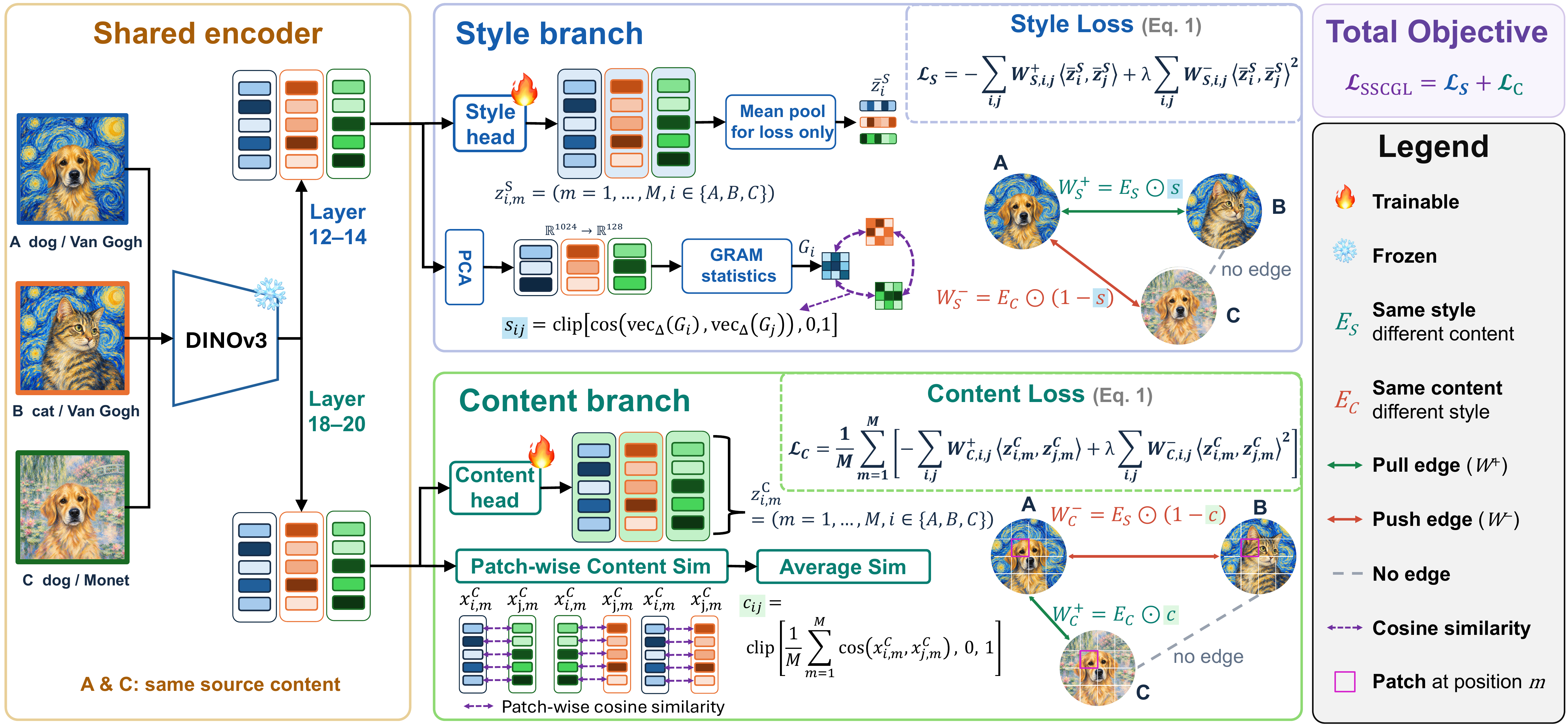}
    \caption{\textbf{Stage 1. ChameleonEncoder training with Signed Style-Content Graph Learning (SSCGL).}
     Given a shared DINOv3 backbone, SSCGL constructs signed, weighted graphs from style and content relations to learn disentangled representations.}
    \vspace{-6mm}
    \label{fig:fig3}
\end{figure}
\textbf{Signed Style-Content Graph Learning (SSCGL).}
To achieve such disentanglement, contrastive learning has been explored for general disentangled representation learning~\citep{matthes2023towards}. In standard contrastive learning, methods such as SimCLR~\citep{chen2020simple} and hard-negative variants such as HCL~\citep{robinson2020contrastive} learn representations invariant to augmentations that are assumed to preserve semantic content by treating differently augmented versions of the same image as positive pairs under InfoNCE-style objectives~\citep{oord2018representation}. However, this augmentation-based supervision is ill-suited to style-content disentanglement, since transformations such as color jittering perturb style-related factors (e.g., brightness, contrast, and color) while the resulting samples are still treated as equivalent.

\begin{wrapfigure}{r}{0.55\linewidth}
    \centering
    \includegraphics[width=\linewidth]{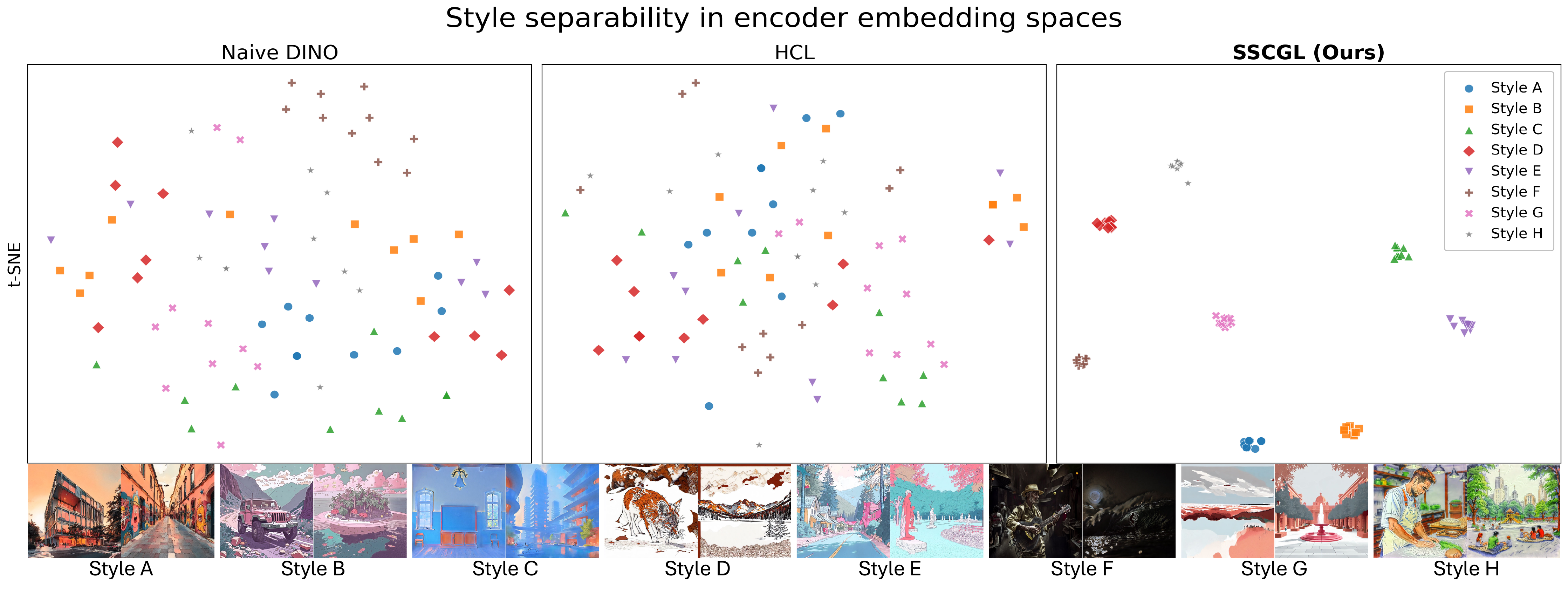}
    \vspace{-8mm}
    \caption{
    \textbf{Style disentanglement in encoder embeddings.}
    t-SNE of 8 styles with 10 samples each.
    Naive DINO and HCL yield entangled style representations, while SSCGL forms clearly separated style clusters.}
    \vspace{-5mm}
    \label{fig:fig5}
\end{wrapfigure} 
\noindent
Hence, rather than constructing positive pairs through augmentation, we leverage explicit style and content annotations available in the training data~\citep{wang2025omnistyle} to reorganize the samples into factor-specific relation sets. This allows us to identify image pairs that share style but differ in content, or share content but differ in style. Crucially, SSCGL applies complementary constraints to the \emph{same image pair} across both representation spaces, pulling it together in one space and pushing it apart in the other. For example, in Fig.~\ref{fig:fig3}, the dog/Gogh and cat/Gogh images share style but differ in content, and are therefore pulled together in the style space while being pushed apart in the content space. Conversely, dog/Gogh and dog/Monet share content but differ in style, yielding the opposite assignment. Using \emph{different image pairs} across the two spaces would break this complementary coupling, allowing each representation to rely on cues from the non-target factor. For example, when aligning dog/Gogh and cat/Gogh in the style space, the model could exploit shared animal-level semantic cues rather than relying solely on their common Gogh style. 
Concretely, we define two relation-specific edge sets: $E_S$ for image pairs that share style but differ in content, and $E_C$ for pairs that share content but differ in style. Unlike conventional contrastive learning that samples isolated anchor-positive pairs, SSCGL includes all within-batch relations satisfying each factor condition, providing group-level relational supervision that encourages factor-wise clustering (Fig.~\ref{fig:fig5}). 
However, these label-defined relations are binary ($0/1$), while the degree of style or content similarity can differ across image pairs. We therefore assign factor-specific weights using frozen DINOv3 features. Specifically, let $s=[s_{ij}]$ denote style similarity measured using Gram statistics, and $c=[c_{ij}]$ denote content similarity measured using patch-wise cosine similarity between corresponding tokens. Combining $E_S$ and $E_C$ with these factor-specific similarities gives $W_S^{+}=E_S\odot s$ and $W_S^{-}=E_C\odot(1-s)$ for the style space, and $W_C^{+}=E_C\odot c$ and $W_C^{-}=E_S\odot(1-c)$ for the content space. Further details in Appendix~\ref{sec:sscgl_theory}.
Given the signed and weighted relations defined above, we jointly optimize the style and content projection heads of \emph{ChameleonEncoder} with the following objectives:

\vspace{-7mm}
\begin{equation}
\label{eq:sscgl}
\resizebox{0.94\linewidth}{!}{$\displaystyle
\mathcal{L}_{\mathrm{S}}
=
\underbrace{-\sum_{i,j} W_{\mathrm{S},ij}^{+}
\langle \bar{z}_{i}^{\mathrm{S}},\bar{z}_{j}^{\mathrm{S}}\rangle}_{\text{positive-edge}}
+\underbrace{\lambda\sum_{i,j} W_{\mathrm{S},ij}^{-}
\langle \bar{z}_{i}^{\mathrm{S}},\bar{z}_{j}^{\mathrm{S}}\rangle^{2}}_{\text{negative-edge}},
\quad
\mathcal{L}_{\mathrm{C}}
=
\frac{1}{M}\sum_{m=1}^{M}
\Big[
-\sum_{i,j} W_{\mathrm{C},ij}^{+}
\langle z_{i,m}^{\mathrm{C}},z_{j,m}^{\mathrm{C}}\rangle
+\lambda\sum_{i,j} W_{\mathrm{C},ij}^{-}
\langle z_{i,m}^{\mathrm{C}},z_{j,m}^{\mathrm{C}}\rangle^{2}
\Big].
$}
\end{equation}
\vspace{-6mm}

where $\bar{z}_i^{\mathrm{S}}$ is the average-pooled style representation, following prior work~\citep{gatys2016image}, while $z_{i,m}^{\mathrm{C}}$ denotes the $m$-th content patch representation, preserving spatial correspondence. $M$ is the number of patch tokens, $\langle\cdot,\cdot\rangle$ the Euclidean inner product, and $\lambda$ balances the positive- and negative-edge terms. We fix $\lambda=1$ for all experiments. We constrain the projected representations to the unit ball and normalize each edge-weight matrix to unit sum. The final objective is $\mathcal{L}_{\mathrm{SSCGL}}=\mathcal{L}_{\mathrm{S}}+\mathcal{L}_{\mathrm{C}}$.

This objective explicitly drives negative-edge pairs toward orthogonality while aligning positive-edge pairs, rather than relying on relative similarity as in InfoNCE. Its global minimum yields style and content representations invariant to content and style, respectively (Appendix~\ref{sec:sscgl_theory}). These factor-selective representations capture the background style from $I_b$ without content leakage while preserving foreground identity from $I_f$, for effective cross-domain compositing.
\begin{figure}[t]
    \centering
    \includegraphics[width=\linewidth]{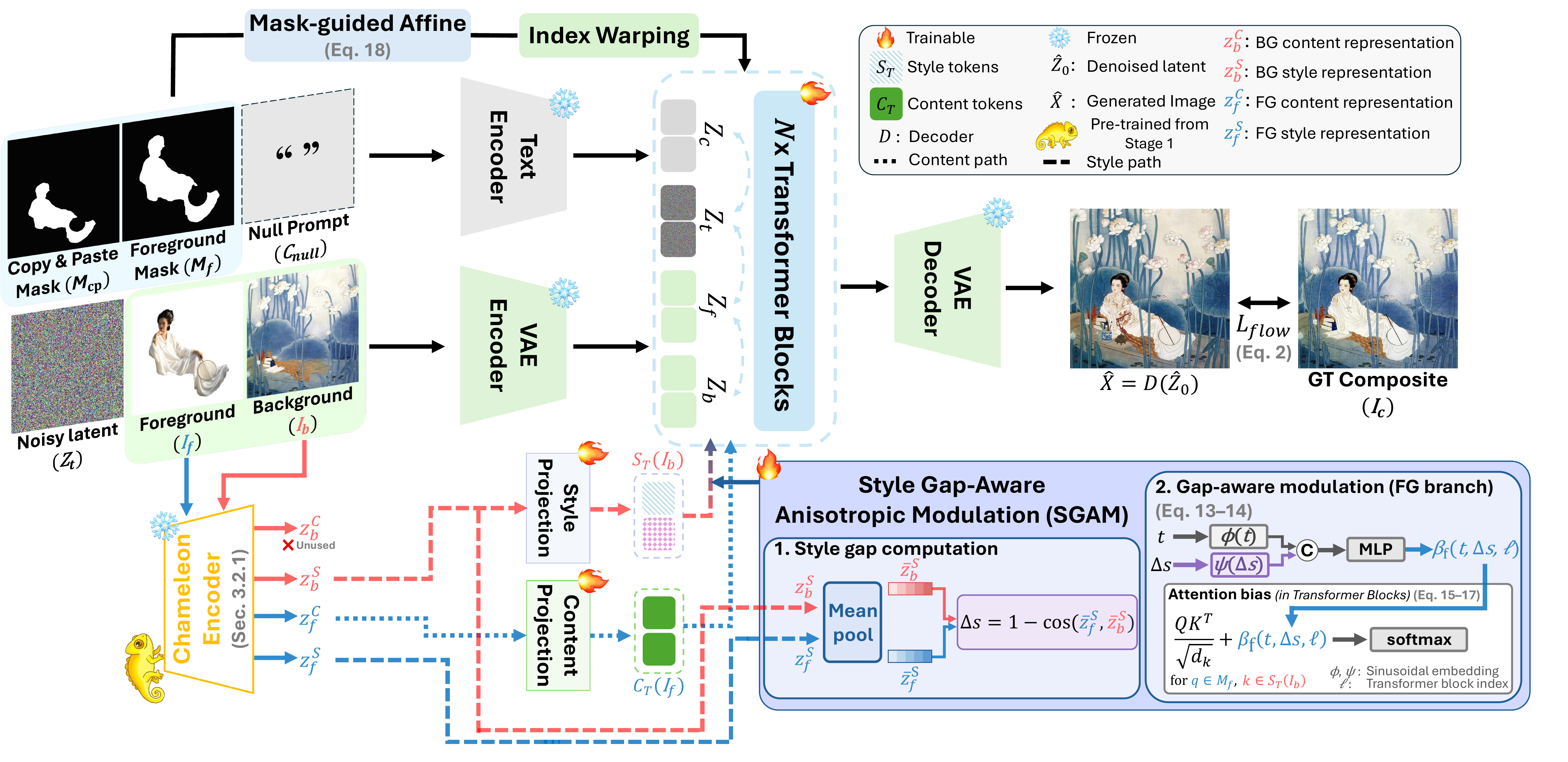}
    \vspace{-8mm}
    \caption{\textbf{Stage 2.} Cross-domain compositing with foreground content and background style tokens from ChameleonEncoder, with SGAM adaptively regulating style injection.}
    \label{fig:fig6}
    \vspace{-5mm}
\end{figure}
\vspace{-2mm}
\subsection{Stage 2 -- Cross-Domain Compositing Model}
\label{sec:stage_generation}
\vspace{-2.5mm}
Building on the disentangled representations learned by ChameleonEncoder in Stage~1 (Sec.~\ref{sec:stage1}), our Stage~2 cross-domain compositing model uses these representations as explicit conditions for the DiT (Fig.~\ref{fig:fig6}). Specifically, ChameleonEncoder extracts content tokens $C_T(I_f)$ from the foreground image to preserve object identity and style tokens $S_T(I_b)$ from the background image to capture the target style, both of which are injected into the DiT. The model is trained on triplets $\{I_f, I_b, I_c\}$ from ChameleonDataset (Sec.~\ref{sec:dataset}), where $I_c$ provides real-image supervision during training and $Z_t$ is initialized from Gaussian noise at inference. The foreground $I_f$ and background $I_b$ are encoded into latents $Z_f$ and $Z_b$, concatenated with $Z_t$, and fed into the DiT. We adopt a null prompt $C_{\text{null}}$ since text is ambiguous for instance-level conditioning, relying instead fully on $I_f$ and $I_b$. For spatial placement, rather than using $M_f$ as a condition token, we exploit $M_f$ and the copy-and-paste mask $M_{cp}$ to compute a positional affine transformation that warps foreground latent indices to their target placement (Appendix~\ref{app:inference_robustness}). To adaptively regulate the influence of the background style tokens $S_T(I_b)$, we propose Style Gap-Aware Anisotropic Modulation (SGAM), detailed in the next subsection.

\noindent\textbf{Style Gap-Aware Anisotropic Modulation (SGAM).}
To effectively harmonize heterogeneous domains between the foreground and background images, the background style tokens $S_T(I_b)$ must be properly injected into the DiT. Since the foreground must adapt to the background style while preserving its identity, excessive style injection may degrade foreground identity, whereas insufficient injection may leave the foreground stylistically inconsistent with the background. Effective style injection should therefore account for \textit{where}, \textit{when}, and \textit{how much} style is injected. Spatially, we focus style injection on foreground tokens, since the background already represents the target style and only the foreground requires adaptation. Moreover, diffusion model inference evolves from noise to structured representations over timesteps, making style injection at early noisy stages less effective as meaningful stylization should occur when semantic structure begins to emerge~\citep{hu2024diffusest}. The appropriate injection strength should also vary with the style gap between $I_f$ and $I_b$. Stylistically similar pairs require only modest adaptation, whereas larger style gaps call for stronger modulation.
While related diffusion-based approaches~\citep{hu2024diffusest,jeong2025structure} modulate conditional information only over diffusion timesteps using \textit{fixed} schedules, we instead introduce Style Gap-Aware Anisotropic Modulation (SGAM), which \textit{adaptively} and anisotropically modulates style injection across three axes: the style gap, spatial region, and diffusion timestep.
Specifically, we map the diffusion timestep $t$ to a sinusoidal time embedding and additionally condition the foreground modulation branch on the foreground-background style gap derived from the ChameleonEncoder style head in Stage~1, while the background branch remains conditioned only on $t$. The resulting modulation coefficients are assigned to query tokens based on their spatial locations via the foreground mask and applied as an attention bias exclusively to keys corresponding to $S_T(I_b)$ at every transformer block (full derivation in \textit{Appendix}).
Consistent with this design, SGAM causes foreground-region queries to attend strongly to $S_T(I_b)$ while adaptively varying the modulation strength according to the style gap and diffusion timestep. We further analyze its spatial, style-gap-aware, timestep-adaptive, and block-wise behaviors in Fig.~\ref{fig:fig8} in \textit{Appendix}.

\vspace{-4mm}
\section{Experiments}
\label{sec:experiments}  
\vspace{-3mm}
\subsection{Experimental Set-up}
\label{sec:Experimental Setups}
\vspace{-2mm}
\noindent \textbf{Evaluation Metrics.}
Following~\citet{lu2023tf, li2025aicomposer}, we measure style consistency with \textbf{CSD}~\citep{somepalli2024investigating}, and foreground preservation with \textbf{LPIPS}~\citep{zhang2018unreasonable} and \textbf{CLIP-I}~\citep{hessel2021clipscore}. We further report a \textbf{VLM-based score}~\citep{ku2024viescore,peng2025dreambench} adapted from~\citet{ju2026editverse} for cross-domain compositing. \noindent \textbf{Benchmarks.} Cross-domain compositing lacks a dedicated benchmark. Existing benchmarks~\citep{lu2023tf, li2025aicomposer} have limited diversity in foregrounds and background styles. We therefore introduce \textbf{ChameleonDataset$_{\text{ev}}$}, covering diverse styles and challenging cases (\textit{e.g.}, reflections and pixel art), and evaluate on both. \textbf{Implementation.} Stage 1 trains ChameleonEncoder on a frozen DINOv3 ViT-L/16. Stage 2 fine-tunes a DiT with LoRA~\citep{hu2022lora}. Full details in Appendix~\ref{app:impl}.
\vspace{-2.5mm}
\begin{figure}[t]
    \vspace{-6pt}
    \centering
    \includegraphics[width=0.935\linewidth]{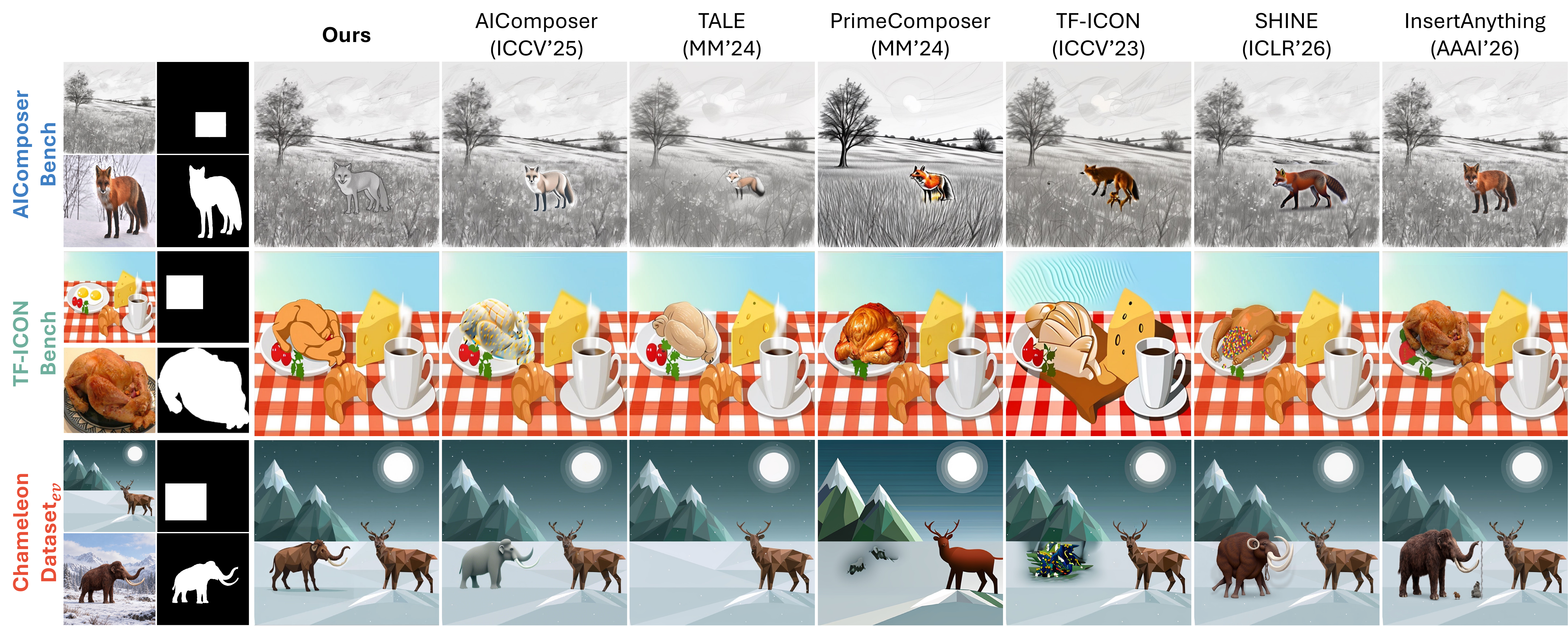}
    \caption{\textbf{Qualitative comparison across three diverse cross-domain compositing benchmarks.}}
    \label{fig:fig7}
\end{figure}
\begin{table*}[t]
\centering
\caption{\textbf{Quantitative comparison on TF-ICON and AIComposer benchmarks.} Reference-based metrics (LPIPS, CLIP-I, CSD) are reported per benchmark, while VLM-based scores (Identity, Style, Composition, Avg\_total) are averaged over the two benchmarks. \textbf{Bold} and \underline{underline} denote the best and second-best results. Each method is evaluated at its backbone's native resolution.}
\label{tab:combined_bench}
\resizebox{\textwidth}{!}{%
\begin{tabular}{l|c|ccc|ccc|cccc}
\toprule
& & \multicolumn{3}{c|}{TF-ICON} & \multicolumn{3}{c|}{AIComposer} & \multicolumn{4}{c}{VLM (avg.)} \\
\cmidrule(lr){3-5} \cmidrule(lr){6-8} \cmidrule(lr){9-12}
Method & Res & LPIPS $\downarrow$ & CLIP-I $\uparrow$ & CSD $\uparrow$ & LPIPS $\downarrow$ & CLIP-I $\uparrow$ & CSD $\uparrow$ & Identity $\uparrow$ & Style $\uparrow$ & Composition $\uparrow$ & Avg\_total $\uparrow$ \\
\midrule
\multicolumn{12}{l}{\textit{In-domain compositing}} \\
\midrule
BLD~(SIGGRAPH'23)              & 512  & 0.7540 & 0.7331 & 0.4619 & 0.6821 & 0.7905 & 0.3624 & 2.07 & 2.71 & 2.08 & 6.86 \\
Paint by Example~(CVPR'23)     & 512  & 0.7501 & 0.7658 & 0.3138 & 0.6754 & 0.8042 & 0.3287 & 2.19 & 2.45 & 2.14 & 6.78 \\
Insert Anything~(AAAI'26)      & 1024 & 0.7281 & 0.7719 & 0.3821 & 0.5418 & 0.8487 & 0.3956 & 2.33 & 2.56 & 2.20 & 7.09 \\
Dreamfuse~(ICCV'25)            & 1024 & 0.6892 & 0.7894 & 0.4035 & 0.5290 & \underline{0.8603} & 0.3941 & 2.36 & 2.51 & 2.07 & 6.94 \\
SHINE~(ICLR'26)                & 512  & 0.7654 & 0.7155 & 0.4362 & 0.6892 & 0.8214 & 0.3387 & 1.97 & 2.62 & 2.10 & 6.69 \\
SHINE~(ICLR'26)                & 1024 & 0.7737 & 0.7165 & 0.4627 & 0.6513 & 0.8377 & 0.3512 & 2.04 & 2.56 & 2.11 & 6.70 \\
\midrule
\multicolumn{12}{l}{\textit{Two-stage compositing}} \\
\midrule
\makecell[l]{StyleSSP~(CVPR'25) \\ + Insert Anything~(AAAI'26)} & 1024 & 0.8410 & 0.7223 & \textbf{0.5535} & 0.4682 & 0.8438 & \textbf{0.4898} & 2.31 & 2.85 & 2.01 & 7.20 \\
\midrule
\multicolumn{12}{l}{\textit{Cross-domain compositing}} \\
\midrule
TF-ICON~(ICCV'23)              & 512  & 0.7740 & 0.7551 & 0.4103 & 0.6640 & 0.7493 & 0.3536 & 1.88 & 2.61 & 1.97 & 6.46 \\
TF-ICON~(ICCV'23)              & 768  & 0.7440 & 0.7581 & 0.4073 & 0.6539 & 0.7700 & 0.3464 & 1.95 & 2.56 & 2.02 & 6.52 \\
TALE~(MM'24)                   & 512  & 0.7990 & 0.5395 & 0.4982 & 0.6164 & 0.7830 & 0.4681 & 1.91 & 2.84 & 2.14 & 6.89 \\
PrimeComposer~(MM'24)          & 512  & 0.7762 & 0.7510 & 0.4285 & 0.6438 & 0.7682 & 0.4124 & 2.10 & 2.71 & 1.95 & 6.78 \\
AIComposer~(ICCV'25)           & 512  & 0.5025 & 0.7946 & 0.4674 & 0.4895 & 0.8412 & 0.4729 & 2.21 & 2.80 & 2.04 & 7.15 \\
AIComposer~(ICCV'25)           & 1024 & \underline{0.4966} & \underline{0.8264} & 0.4826 & \underline{0.4648} & 0.8575 & 0.4848 & 2.40 & 2.84 & 2.09 & 7.33 \\
\midrule
\textbf{Ours}                  & 512  & 0.5018 & 0.8001 & 0.4728 & 0.4724 & 0.8507 & 0.4795 & \underline{2.46} & \underline{2.89} & \underline{2.26} & \underline{7.62} \\
\textbf{Ours}                  & 1024 & \textbf{0.4855} & \textbf{0.8296} & \underline{0.4994} & \textbf{0.4556} & \textbf{0.8637} & \underline{0.4886} & \textbf{2.53} & \textbf{2.90} & \textbf{2.31} & \textbf{7.73} \\
\bottomrule
\end{tabular}%
}
\vspace{-4mm}
\end{table*}

\begin{table}[t]
    \centering
    \begin{minipage}[t]{0.48\linewidth}
        \centering
        \vspace{-2mm}
        \caption{Ablation on the AIComposer benchmark, cumulatively adding SSCGL and SGAM.}
        \label{tab:ablation}
        \renewcommand{\arraystretch}{1.15}
        \footnotesize

        \resizebox{\linewidth}{!}{%
        \begin{tabular}{l cccc}
            \toprule
            Method & LPIPS $\downarrow$ & CLIP-I $\uparrow$ & CSD $\uparrow$ & Aes $\uparrow$ \\
            \midrule
            Baseline
            & 0.4869 & 0.8495 & 0.3992 & 6.8267 \\
            + SSCGL
            & 0.4673 & \textbf{0.8645} & 0.4527 & 6.9209 \\
            + SSCGL + SGAM
            & \textbf{0.4556} & 0.8637 & \textbf{0.4886} & \textbf{7.0296} \\
            \bottomrule
        \vspace{-20mm}    
        \end{tabular}%
        }
    \end{minipage}
    \hfill
    \begin{minipage}[t]{0.50\linewidth}
        \centering
        \vspace{-2mm}
        \caption{User study win rates (\%) across 30 ChameleonDataset$_{\text{ev}}$ cases with 50 participants.}
        \label{tab:userstudy}
        \renewcommand{\arraystretch}{1.15}
        \footnotesize

        \resizebox{\linewidth}{!}{%
        \begin{tabular}{l ccccc}
            \toprule
            Criteria & TF-ICON & TALE & PrimeComposer & AI-Composer & \textbf{Ours} \\
            \midrule
            Identity ($\uparrow$) & 1.8 & 0.1 & 2.3 & 31.4 & \textbf{64.3} \\
            Style ($\uparrow$)    & 1.8 & 0.8 & 1.3 & 25.9 & \textbf{70.2} \\
            Overall ($\uparrow$)  & 2.7 & 0.5 & 1.4 & 25.3 & \textbf{70.1} \\
            \bottomrule
        \end{tabular}%
        }
    \end{minipage}
    \vspace{-3mm}
\end{table}
\subsection{Experimental Results}
\label{sec:comparison}
\vspace{-2mm}
\textbf{Qualitative Comparison.} In Fig.~\ref{fig:fig7}, we compare our method with cross-domain methods~\citep{li2025aicomposer,pham2024tale,wang2024primecomposer,lu2023tf} and in-domain methods~\citep{lu2026does,song2026insert} (two columns on the right). In the last row, the in-domain methods preserve foreground identity but retain a photorealistic appearance, failing to adapt to the low-poly background style. Most cross-domain methods substantially distort foreground identity. AIComposer better preserves foreground identity, yet mainly blends surrounding color tones without reproducing the target style or compositional effects. In contrast, our result preserves identity while matching the low-poly style and producing a plausible grounded shadow. More results are in \textit{Appendix}.

\textbf{Quantitative Comparison.} Quantitative results on TF-ICON and AIComposer benchmarks reported in Tab.~\ref{tab:combined_bench}. In-domain methods generally exhibit low scores on CSD, which measures style consistency, while cross-domain methods tend to achieve lower semantic alignment scores, such as CLIP-I. 2-stage cascaded pipeline with a stylization method and a compositing method~\citep{song2026insert} results in high CSD due to over-stylization (see Fig.~\ref{fig:fig1} (g)). In contrast, our method simultaneously achieves good identity preservation and consistent stylization with best or second-best results across all metrics, which is further supported by a user study on ChameleonDataset$_{\mathrm{ev}}$ (Tab.~\ref{tab:userstudy}). Full results for each dataset including additional metrics (CLIP-T~\citep{radford2021learning} and Aes~\citep{schuhmann2022aesthetic}) are provided in \textit{Appendix}.

\textbf{Ablation Study.}
Tab.~\ref{tab:ablation} presents a component-wise analysis of our method. We use naive LoRA fine-tuning of the DiT on ChameleonDataset$_{\mathrm{tr}}$ as the baseline. Although this baseline produces plausible composites, it fails to fully capture the target style, resulting in relatively lower CSD scores. Replacing the training objective with SSCGL, which disentangles content and style representations for separate injection, consistently improves both CLIP-I and CSD, validating the benefit of explicit representation disentanglement. Adding SGAM further enhances stylization while maintaining stable CLIP-I scores and improving LPIPS, indicating preserved content fidelity. More detailed ablation studies and analyses of individual components are provided in \textit{Appendix}.

\vspace{-3mm}
\section{Conclusion}
\label{sec:conclusion}  
\vspace{-3mm}
We address the challenging problem of cross-domain compositing, which requires preserving foreground identity while adapting to background style. We propose Chameleon, a two-stage training-based framework built on ChameleonDataset, the first large-scale dataset for this task. Our SSCGL objective effectively disentangles style and content, and SGAM adaptively regulates style injection into the DiT. Unlike prior training-free blending-based approaches, our method achieves both strong identity preservation and background-consistent stylization.

\clearpage
\bibliographystyle{iclr2027_conference}
\bibliography{main}
\clearpage

\appendix
\etocdepthtag.toc{mtappendix}
\etocsettagdepth{mtchapter}{none}
\etocsettagdepth{mtappendix}{subsection}
\etocsettocstyle{\noindent{\large\textsc{Appendix Contents}}\par\vspace{1.5em}}{}
\begingroup
\setlength{\parskip}{3.5pt}
\hypersetup{linkcolor=black}
\tableofcontents
\endgroup
\clearpage

\section*{Overview}
In this supplementary material, Sec.~\ref{app:dataset} describes \emph{ChameleonDataset}. 
We revisit previous construction paradigms (Sec.~\ref{app:previous:forward}), 
provide evidence of the synthetic distribution shift they induce (Sec.~\ref{sec:dist_shift}), 
and present the specification and stage-wise construction statistics of 
ChameleonDataset$_{\mathrm{tr}}$ (Sec.~\ref{sec:dataset_statistics}) as well as 
example samples from ChameleonDataset$_{\mathrm{ev}}$ (Sec.~\ref{chameleon_ev}). 
Sec.~\ref{app:method} presents preliminaries on the Diffusion Transformer 
(Sec.~\ref{app:method:dit}), a theoretical analysis of Signed Style-Content Graph Learning 
(SSCGL, Sec.~\ref{sec:sscgl_theory}), the full formulation of Style Gap-Aware Anisotropic 
Modulation (SGAM, Sec.~\ref{sgam_full}), and the batch construction procedure for SSCGL 
(Sec.~\ref{batch}). 
Sec.~\ref{app:impl} reports training details, compute and efficiency analyses 
(Sec.~\ref{compute}), external models and licenses (Sec.~\ref{license}), and the user study 
protocol (Sec.~\ref{user_study}). 
Sec.~\ref{app:abl} provides ablation studies on the encoder design 
(Sec.~\ref{app:abl:encoder}) and SGAM (Sec.~\ref{app:abl:stag}), followed by a qualitative 
component-wise ablation (Sec.~\ref{app:qual_ablation}). 
Sec.~\ref{app:inference_robustness} clarifies the inference assumptions and evaluates 
robustness to coarse localization and imperfect foreground inputs. 
Sec.~\ref{app:quant} presents full quantitative comparisons against cross-domain 
compositing, painterly harmonization, and general-purpose image editing methods 
(Sec.~\ref{app:quant_full}), along with a pose adaptation evaluation under substantial 
viewpoint changes (Sec.~\ref{app:pose_adaptation}). 
Sec.~\ref{app:qual} shows extended benchmark comparisons and results on challenging and 
generalized compositing scenarios (Sec.~\ref{app:qual:challenging}). 
Sec.~\ref{appendix:vlm-prompt} provides the VLM prompts used for evaluation and data 
filtering (Sec.~\ref{sec:vlm_filtering}). 
Finally,

\section{Dataset Details: ChameleonDataset}
\label{app:dataset}

\subsection{Previous Construction Paradigms}
\label{app:previous:forward}
\begin{figure}[H]
    \centering
    \includegraphics[width=\linewidth]{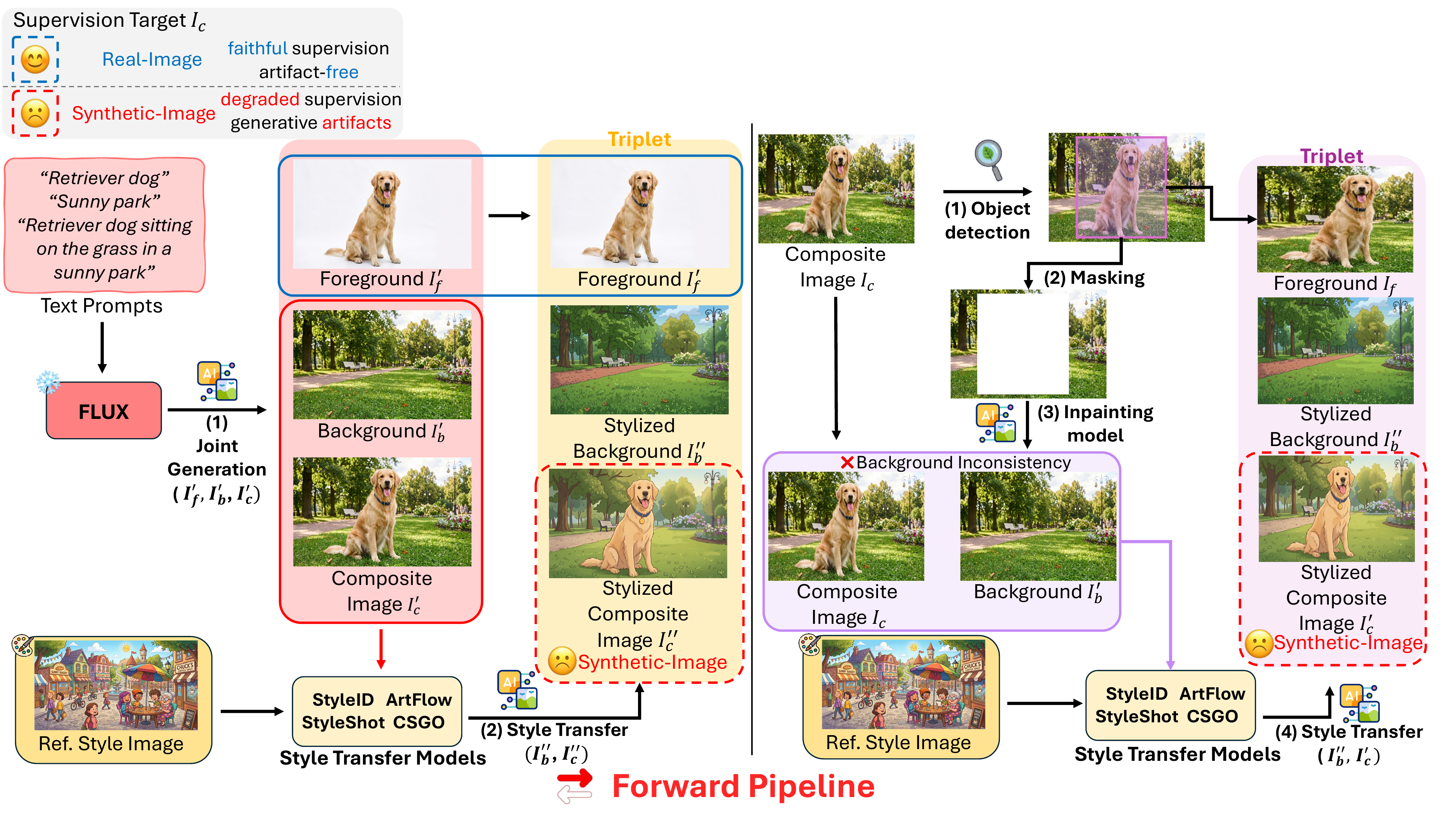}
    \vspace{-25pt}
    \caption{
    \textbf{Forward pipeline.} Prior approaches construct triplets through sequential generation and stylization, producing a synthetic-image supervision target $I_c'$ or $I_c''$, where each $'$ denotes a stage of \textit{synthetic degradation}.}
    \label{fig:fig10}
\end{figure}

\textbf{Forward pipeline}~(top of Fig.~\ref{fig:fig10}) instantiates the synthetic construction through a sequential pipeline that couples compositing and stylization. It admits two entry points. In the first variant~(left in Fig.~\ref{fig:fig10}), a compositing model~\citep{huang2025dreamfuse} jointly denoises the foreground $I_f'$, background $I_b'$, and composite $I_c'$ using a text-to-image backbone~\citep{blackforestlabs2024flux}, so that generative artifacts already arise in $I_c'$. In the second variant~(top-right in Fig.~\ref{fig:fig10}), the pipeline starts from a real composite $I_c$~\citep{yang2023paint}. A foreground $I_f$ is extracted by object detection~\citep{kirillov2023segment}, and the corresponding region is filled by an inpainting model~\citep{suvorov2022resolution} to obtain a synthetic background $I_b'$ for the subsequent stylization stage. A stylization model~\citep{chung2024style, an2021artflow, gao2025styleshot, xing2025csgo} then transforms the input, conditioned on a reference style image, into a stylized composite and a stylized background, yielding $I_c'$ and $I_b'$ in the first variant, and $I_c''$ and $I_b''$ in the second. The fundamental issue is that the stylized composite, whether $I_c'$ or $I_c''$, is a \emph{synthetic-image}, inheriting artifacts accumulated across both stages and ultimately following a synthetic-image distribution. Using such a composite as degraded supervision trains the model to reproduce this distribution rather than the real-image distribution, yielding an inherently suboptimal mapping regardless of model capacity.

\subsection{Evidence of Synthetic Distribution Shift}
\label{sec:dist_shift}
\begin{figure}[H]
    \vspace{-10pt}
    \centering
    \includegraphics[width=0.85\linewidth]{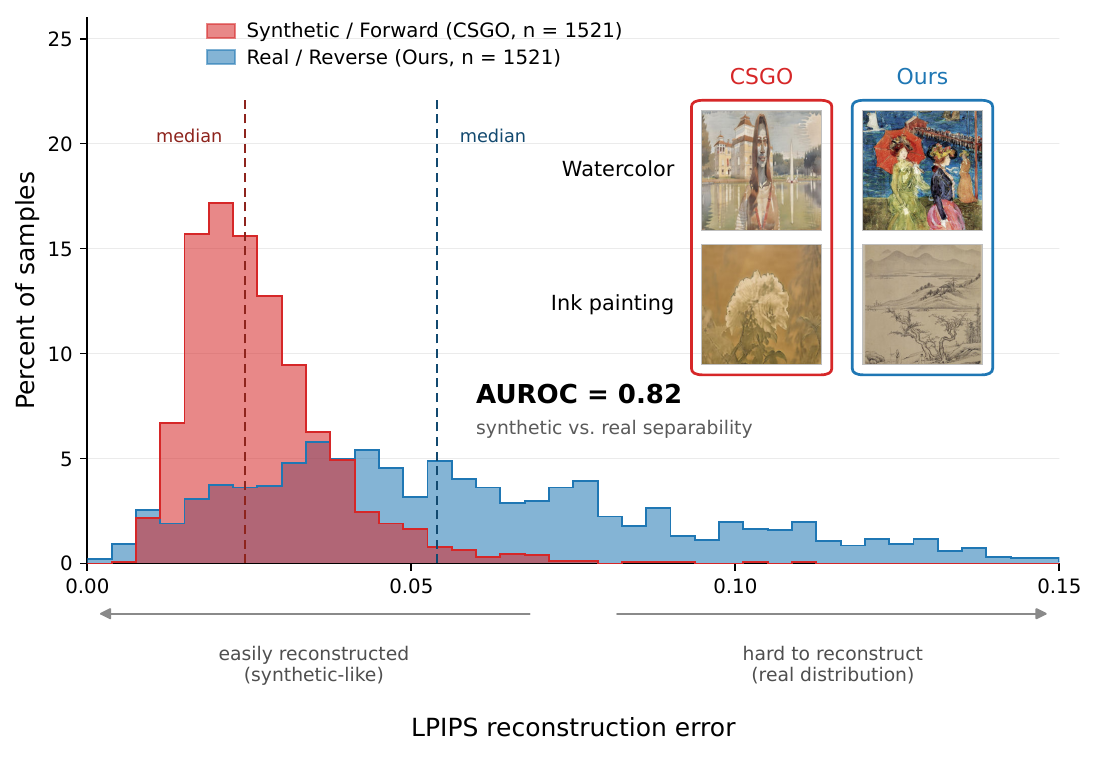}
    \vspace{-10pt}
    \caption{
    \textbf{Forward--reverse reconstruction gap.}
    We compare AeroBLADE-based LPIPS reconstruction errors of stylized composites from our reverse pipeline (Real / Reverse, Ours) and a forward pipeline using CSGO for stylization (Synthetic / Forward, CSGO). Both sets contain 1,521 samples with matched style and object distributions, isolating the effect of synthetic generation. Forward composites exhibit lower errors, a characteristic signature of diffusion-generated images, whereas reverse composites show higher errors, deviating from this synthetic-image signature. The resulting AUROC of 0.82 indicates clear separability between the two pipelines.}
    \label{fig:fig_add2} 
\end{figure}

AeroBLADE~\citep{ricker2024aeroblade} detects generated images from latent diffusion models using autoencoder reconstruction error, where lower reconstruction errors indicate a stronger generated-image signature.
Given an image $x$, the method reconstructs it as $\hat{x} = \mathcal{D}(\mathcal{E}(x))$, where $\mathcal{E}$ and $\mathcal{D}$ denote the encoder and decoder of the latent diffusion model's autoencoder, respectively.
The reconstruction error is then defined as $e(x) = \mathrm{LPIPS}(x, \hat{x})$, the perceptual distance between $x$ and $\hat{x}$.
We repurpose this signal to compare the synthetic artifacts introduced by the reverse (Fig.~\ref{fig:fig2} of the main paper) and forward (Fig.~\ref{fig:fig10}) generation paths. Specifically, we generate data through each pipeline and measure $e(x)$ on the resulting stylized composites: $I_c$ (real) from the reverse path and $I_c'$ or $I_c''$ (synthetic) from the forward path. Each set contains 1,521 samples matched by style and object category.
To quantify the separation between the two reconstruction-error distributions, we report the area under the receiver operating characteristic curve (AUROC)~\citep{fawcett2006introduction}, which in our setting corresponds to the probability that a randomly selected reverse-pipeline composite yields a higher $e(x)$ than a randomly selected forward-pipeline composite. As shown in Fig.~\ref{fig:fig_add2}, the two pipelines produce clearly distinct reconstruction-error distributions, with an AUROC of 0.82. In other words, in 82\% of randomly sampled reverse--forward pairs, the reverse composite exhibits a higher reconstruction error, indicating a weaker generated-image signature.
Interestingly, we find that synthetic and real signatures separate differently across styles.
In Tab.~\ref{tab:stylewise}, the Real/Reverse and Synthetic/Forward columns report the LPIPS reconstruction error $e(x)$, and AUROC is computed in the same manner as above.
This separation is most pronounced for texture-rich styles, where generation leaves clear synthetic artifacts, whereas the gap narrows for line-dominant styles, whose simpler structures leave fewer detectable artifacts.

\begin{table}[H]
    \centering
    \vspace{-25pt}
    \caption{
    \textbf{Style-wise analysis of synthetic generation artifacts.}
    We report the mean reconstruction error $e(x)$ of real composites from our reverse pipeline and synthetic composites from the forward pipeline (CSGO) per style, their ratio, and the AUROC between the two distributions.
    Styles are sorted in descending order of AUROC.
    Texture-rich styles (top rows) exhibit larger discrepancies, whereas line-dominant styles (bottom rows) show relatively small gaps.}
    \label{tab:stylewise}
    \setlength{\tabcolsep}{4pt}
    \renewcommand{\arraystretch}{1.0}
    \begin{tabular}{lcccc}
        \toprule
        \textbf{Style}
        & \makecell{\textbf{Real /}\\\textbf{Reverse (Ours)}}
        & \makecell{\textbf{Synthetic /}\\\textbf{Forward (CSGO)}}
        & \textbf{Ratio}
        & \textbf{AUROC} \\
        \midrule
        Ink painting
        & $0.077 \pm 0.030$
        & $0.025 \pm 0.008$
        & $3.1\times$
        & $0.963$ \\
        Ukiyo-e
        & $0.041 \pm 0.009$
        & $0.024 \pm 0.006$
        & $1.7\times$
        & $0.954$ \\
        Baroque
        & $0.089 \pm 0.037$
        & $0.035 \pm 0.011$
        & $2.5\times$
        & $0.953$ \\
        Game style
        & $0.072 \pm 0.033$
        & $0.027 \pm 0.011$
        & $2.7\times$
        & $0.914$ \\
        Watercolor
        & $0.073 \pm 0.036$
        & $0.032 \pm 0.013$
        & $2.3\times$
        & $0.886$ \\
        Expressionism
        & $0.069 \pm 0.041$
        & $0.032 \pm 0.015$
        & $2.2\times$
        & $0.834$ \\
        Traditional Chinese
        & $0.039 \pm 0.021$
        & $0.022 \pm 0.006$
        & $1.8\times$
        & $0.777$ \\
        Hand-drawn
        & $0.024 \pm 0.019$
        & $0.017 \pm 0.006$
        & $1.4\times$
        & $0.583$ \\
        \bottomrule
    \end{tabular}
\end{table}

\subsection{ChameleonDataset$_{\mathrm{tr}}$} 

Our ChameleonDataset$_{\mathrm{tr}}$ contains diverse foreground object categories (2,000) and stylized backgrounds (1,171), as summarized in Tab.~\ref{tab:dataset_spec}. This diversity is enabled by our reverse pipeline (Fig.~\ref{fig:fig6}), which starts from curated real-image composites $I_c$ rather than relying on multiple style-specific LoRAs or stylization model zoos~\citep{an2021artflow,hu2024diffusest}, which are typically limited to predefined style categories. As a result, the dataset covers a broad range of visual domains, styles, and compositional scenarios, enabling the model to learn diverse cross-domain compositing patterns from data.

\begin{table}[H]
\centering
\footnotesize
\setlength{\tabcolsep}{4pt}
\renewcommand{\arraystretch}{1.08}
\caption{\textbf{ChameleonDataset$_{\mathrm{tr}}$ construction specification.}}
\label{tab:dataset_spec}
\resizebox{\linewidth}{!}{%
\begin{tabular}{l|l|p{0.55\linewidth}}
\toprule
Component & Specification & Details \\
\midrule
Categories & Total categories & $2{,}000$ object categories spanning diverse semantic domains \\
\midrule
Category domains & Main groups &
Human, Animal, Plant, Food, Clothing/Accessories, Household/Object, Vehicle, Building/Structure, Nature/Scene \\
\midrule
Category examples
& Human-related & person, man, woman, child, baby, face, hand, hair, eye \\
& Food-related & fruit, vegetable, cake, bread, pizza, burger, drink \\
& Animal-related & dog, cat, horse, bird, fish, butterfly, elephant \\
& Vehicle-related & car, bus, truck, bicycle, motorcycle, train, airplane \\
& Clothing/Accessory-related & dress, t-shirt, jacket, coat, hat, shoes, bag, necklace \\
\midrule
Styles & Total styles & $1{,}171$ artistic and illustrative styles \\
\midrule
Style domains
& Art movements & impressionism, baroque, surrealism, cubism \\
& Eastern media & woodblock print, ink wash painting \\
& Western media & oil painting, watercolor, pastel painting \\
& Non-painting styles & sculpture, pencil sketch, engraving \\
& Illustrative styles & cartoon, comics, kids' drawing \\
\midrule
Reference augmentation
& Viewpoint & front-right quarter view $(40^\circ)$ and related azimuth variations \\
& Distance & close-up, medium-shot, and wide-shot compositions \\
& Elevation & eye-level, high-angle, and low-angle views \\
& Resolution & multi-resolution generation at $512$, $768$, and $1024$ \\
\midrule
Buckets & Resolution buckets &
$896{\times}1152$, $384{\times}640$, $576{\times}960$, $672{\times}864$, $1024{\times}1024$, $448{\times}576$, $768{\times}1280$, $320{\times}704$, $512{\times}512$, $640{\times}1536$, $768{\times}768$, $480{\times}1056$, $256{\times}768$, $384{\times}1152$, $640{\times}1408$, $320{\times}768$, $480{\times}1152$, $512{\times}1536$ \\
\bottomrule
\end{tabular}%
}
\vspace{-4mm}
\end{table}
\clearpage
\subsection{Dataset Construction Statistics}
\label{sec:dataset_statistics}
Tab.~\ref{tab:dataset_construction_stats} reports stage-wise statistics for our reverse data construction pipeline (Fig.~\ref{fig:fig2}).

\begin{table}[H]
    \centering
    \footnotesize
    \setlength{\tabcolsep}{8pt}
    \renewcommand{\arraystretch}{1.15}
    \caption{
    \textbf{Stage-wise statistics of ChameleonDataset$_{\mathrm{tr}}$ construction.}
    }
    \label{tab:dataset_construction_stats}
    \begin{tabular}{lc}
        \toprule
        Stage & Count \\
        \midrule
        Raw composite images                & 78K  \\
        (1) Object labels                   & 139K \\
        (2a) SAM-kept labels                & 67K  \\
        (2b) Segmented objects              & 226K \\
        (3) Filtered samples                & 185K \\
        (4) Generated references            & 183K \\
        (5) Final samples                   & 199K \\
        \bottomrule
    \end{tabular}
\end{table}

During (1) object label generation, Qwen3-VL proposes up to three candidates per image (e.g., \texttt{``Lion, Book, Desk''}) with confidence above 0.85, yielding 1.79 labels per image on average. This cap prevents scene-level over-representation, as multiple foreground objects from the same $I_c$ share the same scene context and background style. In (2) object segmentation, a SAM 3 threshold of 0.5, chosen from manual inspection of 500 samples, removes 52\% of the labels. The retained labels can correspond to multiple object instances, resulting in 226K segmented object candidates. In (3) data filtering, 82.1\% of the candidates are retained using a keep score $\geq 70$ and recoverability $\geq 0.35$, with the filtering criteria detailed in Appendix~\ref{sec:vlm_filtering}. In (4) reference generation, CLIP-I filtering removes approximately 1\% of generated references with identity changes. Finally, in (5) appearance variation, we perturb camera parameters (e.g., azimuth, elevation) to obtain $I_f''$ with diverse poses and lighting. This encourages the model to adapt the foreground pose and lighting toward those of $I_c$, rather than naively preserving the perturbed input. For instance, when $I_f''$ depicts a lion from a top view while $I_c$ shows it from the side, the supervision encourages adaptation toward the side-view target. This stage increases the dataset by 12\%, after which CLIP-I filtering removes approximately 3\% relative to the original dataset size, resulting in a net increase of 9\%.
\clearpage

\subsection{ChameleonDataset$_{\mathrm{ev}}$}
\label{chameleon_ev}
\begin{figure}[H]
    \centering
    \includegraphics[width=\linewidth]{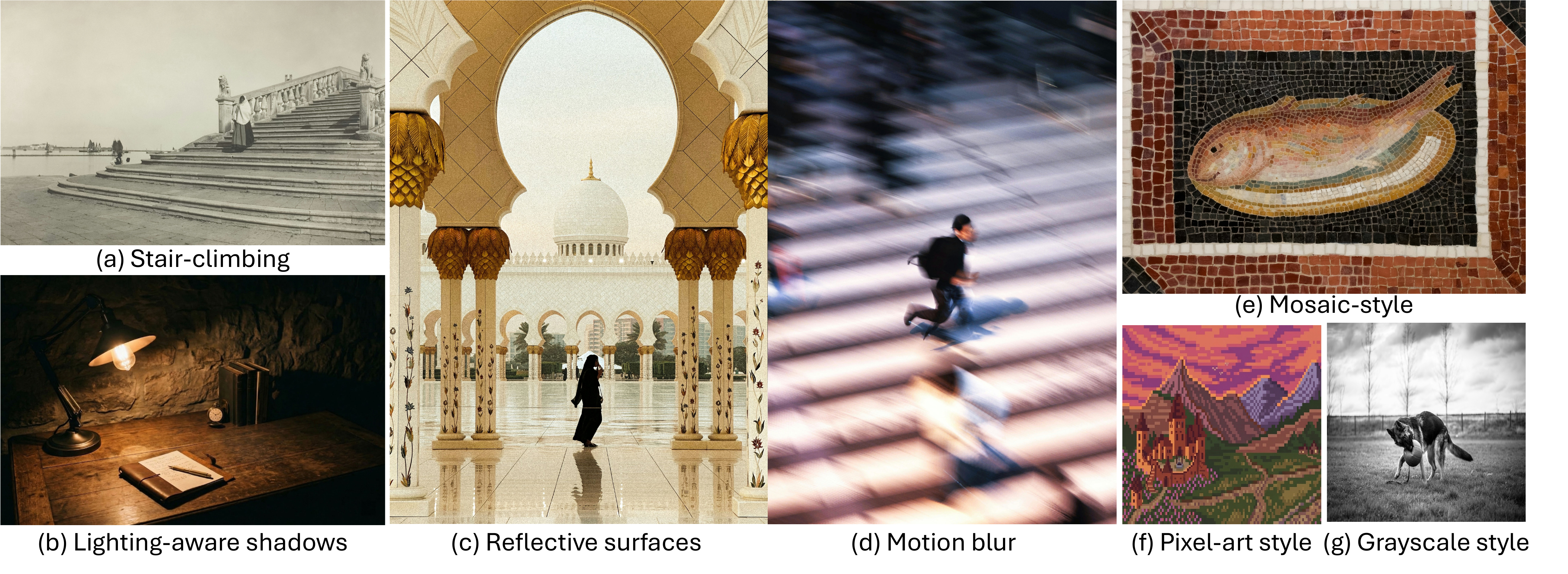}
    \vspace{-20pt}
    \caption{\textbf{Examples of ChameleonDataset$_{\mathrm{ev}}$.} Our benchmark includes challenging compositional scenarios, including (a) stair-climbing, (b) lighting-aware shadows, and (c) reflective surfaces, as well as diverse styles such as (d) motion blur and (e) mosaic-style artwork.}
    \label{fig:eval}
    \vspace{-10pt}
\end{figure}

Existing cross-domain compositing benchmarks provide limited coverage of diverse styles and compositional scenarios. TF-ICON~\citep{lu2023tf} organizes its evaluation data into four coarse domains: real images, paintings, sketches, and cartoons. Although AI-Composer~\citep{li2025aicomposer} includes a broader range of styles, it repeatedly reuses foregrounds and backgrounds, such that a larger number of samples does not necessarily correspond to equally diverse evaluation scenarios. Moreover, as it builds upon TF-ICON and Magic Insert~\citep{ruiz2025magic}, its evaluation data partially overlap with existing benchmarks.
In contrast, ChameleonDataset$_{\mathrm{ev}}$ prioritizes evaluation quality by independently curating diverse, compositionally plausible, and challenging scenarios. Rather than recombining foregrounds and backgrounds from a fixed asset pool, we select the object, background, and insertion placement for each scene, ensuring that each sample represents a meaningful object--scene interaction. 
As shown in Fig.~\ref{fig:eval}, our benchmark evaluates both object--background composition, through scenes such as stair climbing, reflective surfaces, and lighting-aware shadows, and a broad spectrum of visual styles. These include traditional East Asian art, classical and impressionist painting, watercolor and expressionism, pencil and ink drawing, vector graphics, pixel art, 3D rendering, stained glass, mosaics, animation, storybook illustration, and photographic styles. In total, the benchmark spans roughly 25 style families and diverse object categories, including humans, animals, vehicles, buildings, text, food, and plants. Additional foreground objects and stylized backgrounds appear in the qualitative results in Figs.~\ref{fig:fig_add1} , \ref{fig:fig_har}, and \ref{fig:fig_commercial}. All samples are built from publicly available, license-free images~\citep{pixabay,unsplash,pexels}.

\clearpage

\section{Preliminaries, Derivation, and Algorithm}
\label{app:method}
\subsection{Preliminaries: Diffusion Transformer}
\label{app:method:dit}
Our framework (Sec.~\ref{sec:stage_generation}) builds on FLUX.1-dev~\citep{blackforestlabs2024flux}, operating in the latent space. A VAE~\citep{kingma2013auto} encodes an image into a latent $Z_0 \in \mathbb{R}^{H \times W \times C}$, and a T5 encoder~\citep{raffel2020exploring} produces text embeddings $Z_c \in \mathbb{R}^{L \times D}$. The model is trained under a flow matching formulation~\citep{liu2023flow}, which predicts the velocity field along a linear interpolation between Gaussian noise $Z_1 \sim \mathcal{N}(0, I)$ and the target latent $Z_0$. Specifically, we consider the path $Z_t = t Z_0 + (1 - t) Z_1$, where $t \in [0,1]$, and the target velocity $v^\ast = Z_0 - Z_1$. The training objective is:
\begin{equation}
\mathcal{L}_{\text{flow}} = \mathbb{E}_{t, Z_0, Z_1} \left[ \| v_\theta(Z_t, Z_c, t) - v^\ast \|_2^2 \right].
\label{eq:flow}
\end{equation}
For brevity, Fig.~\ref{fig:fig6} depicts $\mathcal{L}_{\text{flow}}$ between the decoded 
output $\hat{X} = D(\hat{Z}_0)$ and $I_c$. In practice, $\mathcal{L}_{\text{flow}}$ is 
computed in the latent space, where the target latent is $Z_0 = \mathcal{E}(I_c)$.

\subsection{Theoretical Analysis of Signed Style-Content Graph Learning}
\label{sec:sscgl_theory}

Under an idealized style-content factorization, we establish that minimizing the SSCGL objective yields factor-selective style and content representations. 

\paragraph{Factor-selective disentanglement.}
Let \(\mathcal{C}\) and \(\mathcal{S}\) denote the sets of content and style
factors, respectively, and denote a sample with content
\(c \in \mathcal{C}\) and style \(s \in \mathcal{S}\) by \(x_{c,s}\), with \(\bar z^{\mathrm S}_{c,s}\) and \(z^{\mathrm C}_{c,s,m}\) denoting its style representation and \(m\)-th patch-wise content representation, respectively.
We consider the full Cartesian product
\(\mathcal{X}=\mathcal{C}\times\mathcal{S}\), where every content factor
is observed with every style factor. As defined in Sec.~\ref{sec:method}, \(E_S\) connects samples with the same style and different content, whereas \(E_C\) connects samples with the same content and different styles. In the
style space, \(E_S\) forms positive edges and \(E_C\) forms negative edges, while the signs are reversed in the content space.

These signed relations are optimized through the SSCGL objective in
Eq.~\eqref{eq:sscgl}. For the style branch,
\[
\mathcal{L}_{\mathrm S}
=
-\sum_{i,j} W_{\mathrm S,ij}^{+}
\left\langle \bar z_i^{\mathrm S}, \bar z_j^{\mathrm S}\right\rangle
+
\lambda
\sum_{i,j} W_{\mathrm S,ij}^{-}
\left\langle \bar z_i^{\mathrm S}, \bar z_j^{\mathrm S}\right\rangle^{2}.
\]
Under the unit-ball constraint in Sec.~\ref{sec:stage1}, the
Cauchy--Schwarz inequality gives
\(\langle \bar z_i^{\mathrm S}, \bar z_j^{\mathrm S}\rangle \leq 1\).
Combined with the unit-sum normalization of \(W_{\mathrm S}^{+}\), this yields
\begin{equation}
-\sum_{i,j} W_{\mathrm S,ij}^{+}
\left\langle \bar z_i^{\mathrm S}, \bar z_j^{\mathrm S}\right\rangle
\geq
-\sum_{i,j} W_{\mathrm S,ij}^{+}
=
-1.
\label{eq:sscgl_positive_bound}
\end{equation}
Since the weighted squared negative-edge term is non-negative,
\begin{equation}
\mathcal{L}_{\mathrm S} \geq -1.
\label{eq:sscgl_style_bound}
\end{equation}
The same argument applies independently to each patch in
\(\mathcal{L}_{\mathrm C}\). Since the content objective averages over
\(M\) patches, we obtain
\begin{equation}
\mathcal{L}_{\mathrm C} \geq -1,
\qquad
\mathcal{L}_{\mathrm{SSCGL}}
=
\mathcal{L}_{\mathrm S}+\mathcal{L}_{\mathrm C}
\geq -2.
\label{eq:sscgl_total_bound}
\end{equation}

\textit{Proposition 1 (Factor-selective optimum).}
Under the idealized full Cartesian setting described above, suppose that
\(\lambda>0\), all active edge weights in
\(W_{\mathrm S}^{\pm}\) and \(W_{\mathrm C}^{\pm}\) are strictly positive,
and the style and content embedding spaces have sufficient dimension to
represent the factor-selective solution.
Then the lower bound
\(\mathcal L_{\mathrm{SSCGL}}=-2\) in
Eq.~\eqref{eq:sscgl_total_bound} is attained if and only if the style and
content representations are factor-selective. Specifically, for each style
\(s\), all samples sharing that style map to a common unit prototype \(u_s\),
independent of content, while distinct style prototypes are pairwise
orthogonal:
\begin{equation}
\bar z^{\mathrm S}_{c,s}=u_s
\quad \text{for all } c\in\mathcal C,
\qquad
\|u_s\|_2=1,
\qquad
\langle u_s,u_{s'}\rangle=0
\quad (s\neq s').
\label{eq:sscgl_style_solution}
\end{equation}
Likewise, for each content \(c\) and patch \(m\), all style variations map
to a common unit prototype \(v_{c,m}\), independent of style, while distinct
content prototypes are pairwise orthogonal:
\begin{equation}
z^{\mathrm C}_{c,s,m}=v_{c,m}
\quad \text{for all } s\in\mathcal S,
\qquad
\|v_{c,m}\|_2=1,
\qquad
\langle v_{c,m},v_{c',m}\rangle=0
\quad (c\neq c').
\label{eq:sscgl_content_solution}
\end{equation}
Consequently, under these assumptions, every global optimum satisfies,
for all \(c,c'\in\mathcal C\), \(s,s'\in\mathcal S\), and each patch \(m\),
\begin{equation}
\bar z^{\mathrm S}_{c,s}
=
\bar z^{\mathrm S}_{c',s},
\qquad
z^{\mathrm C}_{c,s,m}
=
z^{\mathrm C}_{c,s',m}.
\label{eq:sscgl_invariance}
\end{equation}
Thus, the style representation is invariant to content, while the content
representation is invariant to style.
\textit{Proof.}
For the only-if direction, suppose that the lower bound
\(\mathcal L_{\mathrm{SSCGL}}=-2\) is attained.
Since \(\mathcal L_{\mathrm S}\geq-1\) and
\(\mathcal L_{\mathrm C}\geq-1\), this requires
\(\mathcal L_{\mathrm S}=\mathcal L_{\mathrm C}=-1\).
We first consider the style branch. Attaining
\(\mathcal L_{\mathrm S}=-1\) requires the positive-edge term to attain
its lower bound \(-1\) in Eq.~\eqref{eq:sscgl_positive_bound}, while the
non-negative negative-edge term must vanish. Since all active weights in
\(W_{\mathrm S}^{+}\) are strictly positive and sum to one, equality in
Eq.~\eqref{eq:sscgl_positive_bound} requires
\begin{equation}
\left\langle
\bar z_i^{\mathrm S},
\bar z_j^{\mathrm S}
\right\rangle
=1
\qquad
\text{for every active }(i,j)\in E_S.
\end{equation}

Under the unit-ball constraint, this equality is possible only when
\(\bar z_i^{\mathrm S}\) and \(\bar z_j^{\mathrm S}\) are identical unit
vectors. Since \(E_S\) connects samples sharing the same style, all samples
with style \(s\) therefore share a common unit prototype \(u_s\), yielding
\(\bar z^{\mathrm S}_{c,s}=u_s\) for all \(c\in\mathcal C\).

The negative-edge term must simultaneously vanish. Since
\(\lambda>0\) and all active weights in \(W_{\mathrm S}^{-}\) are strictly
positive, this requires
\begin{equation}
\left\langle
\bar z_i^{\mathrm S},
\bar z_j^{\mathrm S}
\right\rangle
=0
\qquad
\text{for every active }(i,j)\in E_C.
\end{equation}
Because \(E_C\) connects samples with the same content and different styles,
the full Cartesian setting implies
\(\langle u_s,u_{s'}\rangle=0\) for all \(s\neq s'\).
Thus, the style branch satisfies the factor-selective structure in
Eq.~\eqref{eq:sscgl_style_solution}.

The content branch follows symmetrically. For each patch \(m\), positive
\(E_C\) edges require
\(\langle z^{\mathrm C}_{i,m}, z^{\mathrm C}_{j,m}\rangle=1\), so all
style variations of the same content share a common unit prototype
\(v_{c,m}\). Likewise, the vanishing negative-edge term on \(E_S\) requires
\(\langle v_{c,m},v_{c',m}\rangle=0\) for \(c\neq c'\).
Thus, Eq.~\eqref{eq:sscgl_content_solution} holds for every patch \(m\),
completing the only-if direction.

For the if direction, suppose that
Eqs.~\eqref{eq:sscgl_style_solution} and
\eqref{eq:sscgl_content_solution} hold.
Under this factor-selective structure, positive edges connect samples that
share the target factor and therefore map to the same unit prototype,
whereas negative edges connect distinct target-factor prototypes, which are
orthogonal. Hence, positive edges have inner product \(1\), while negative
edges have inner product \(0\). By the unit-sum normalization of the positive
edge weights, the style objective and each patch-wise content objective
attain \(-1\), while all negative-edge terms vanish. Therefore,
\(\mathcal L_{\mathrm S}=\mathcal L_{\mathrm C}=-1\) and
\(\mathcal L_{\mathrm{SSCGL}}=-2\).
\hfill\(\square\)

\paragraph{Why factor-specific weighting?}
The result above shows that the signed relation structure determines
\emph{which} factor should be preserved or suppressed. However, a binary
edge sign alone cannot express \emph{how strongly} a pairwise relation should
be enforced. In real data, discrete factor labels capture only coarse
pairwise relations. For example, Expressionism and Fauvism have different style
labels but share visual attributes such as bold colors and expressive
brushwork, whereas Fauvism and pixel art are substantially more distinct.
Assigning identical negative weights to both pairs would therefore impose
the same decorrelation strength despite their different degrees of stylistic
similarity.

SSCGL resolves this by retaining the signed relation while modulating the
strength of each pairwise constraint with factor-specific similarities.
Let \(a_{ij}=\langle \bar z_i^S,\bar z_j^S\rangle\). For a positive style edge,
\begin{equation}
\frac{\partial (-W^+_{S,ij}a_{ij})}{\partial a_{ij}}
=
-W^+_{S,ij}
\propto
-E_{S,ij}s_{ij},
\end{equation}
whereas for a negative style edge,
\begin{equation}
\frac{\partial (\lambda W^-_{S,ij}a_{ij}^{2})}{\partial a_{ij}}
=
2\lambda W^-_{S,ij}a_{ij}
\propto
E_{C,ij}(1-s_{ij})a_{ij}.
\end{equation}
Thus, on positive edges, a larger \(s_{ij}\) places stronger optimization
pressure on increasing \(a_{ij}\), pulling the pair more strongly toward
the maximum similarity, \(a_{ij}=1\). Conversely, on negative edges, a
smaller \(s_{ij}\) increases \(1-s_{ij}\), assigning a larger penalty to
nonzero \(a_{ij}\) and thereby pushing the pair more strongly toward
orthogonality. Hence, \(s\) and \(1-s\) calibrate the relative strengths
of attraction and decorrelation according to the underlying style similarity.
The content weights \(W_C^+\) and \(W_C^-\) play the symmetric role in the
content space.

Taken together, the theoretical analysis above establishes the factor-selective role of the signed graph and clarifies the complementary role of factor-specific weighting. While the signed relations alone characterize the factor-selective solution under the idealized full Cartesian setting, practical settings with finite model capacity and non-ideal data call for finer control over the strength of individual relations. SSCGL therefore uses continuous factor similarities to refine the strength of each pairwise relation. Rather than treating all binary relations equally, the factor-specific weighting adaptively modulates each pairwise constraint according to its factor similarity, providing a more fine-grained learning signal for style-content disentanglement. We further verify this effect empirically in Tab.~\ref{tab:weighting}, where factor-specific weighting improves the disentanglement margin for both style and content over uniform weighting
on the same signed graph.

\subsection{Style Gap-Aware Anisotropic Modulation (SGAM): Full Formulation}
\label{sgam_full}
Specifically, we map the diffusion timestep \(t\) to a sinusoidal embedding \(\phi(t)\). Let \(S_{\mathrm f}\) and \(S_{\mathrm b}\) denote the Stage-1 ChameleonEncoder style representations of the foreground object and background image, respectively. We define the foreground--background style gap as \(\Delta s = 1-\cos(S_{\mathrm f},S_{\mathrm b})\). We then encode \(\Delta s\) using the same sinusoidal encoding scheme as \(t\), yielding \(\psi(\Delta s)\). Since the foreground must adapt toward the background style while the background already represents the target style, the foreground branch is conditioned on both \(\phi(t)\) and \(\psi(\Delta s)\), whereas the background branch remains conditioned only on \(\phi(t)\). The resulting features are linearly projected to obtain layer-wise modulation logits:

\begin{equation}
\begin{aligned}
z_{\mathrm f}(t,\Delta s)
&=
H^{(\mathrm f)}
\,\mathrm{MLP}_{\mathrm f}
\!\left([\phi(t),\,\psi(\Delta s)]\right)
+\beta_0^{(\mathrm f)}
\in \mathbb{R}^L,\\
z_{\mathrm b}(t)
&=
H^{(\mathrm b)}
\,\mathrm{MLP}_{\mathrm b}
\!\left(\phi(t)\right)
+\beta_0^{(\mathrm b)}
\in \mathbb{R}^L,
\end{aligned}
\label{eq:sgam1}
\end{equation}

where \(H^{(\mathrm f)}, H^{(\mathrm b)} \in \mathbb{R}^{L \times h}\) project the \(h\)-dimensional MLP features to the \(L\) transformer blocks, and \(\beta_0^{(\mathrm f)}, \beta_0^{(\mathrm b)} \in \mathbb{R}^{L}\) are learned biases. Accordingly, \(z_{\mathrm f}(t,\Delta s,\ell)\) and \(z_{\mathrm b}(t,\ell)\) denote the foreground and background modulation logits for the \(\ell\)-th block, respectively. These logits are converted to modulation coefficients via
\begin{equation}
\begin{aligned}
\beta_{\mathrm f}(t,\Delta s,\ell)
&= \sigma\bigl(z_{\mathrm f}(t,\Delta s,\ell)\bigr),\\
\beta_{\mathrm b}(t,\ell)
&= \sigma\bigl(z_{\mathrm b}(t,\ell)\bigr),
\end{aligned}
\label{eq:sgam2}
\end{equation}
where \(\sigma(\cdot)\) denotes the sigmoid function. Given a binary foreground mask \(m(q)\in\{0,1\}\), each query token \(q\) is assigned a modulation coefficient according to its spatial location:
\begin{equation}
\gamma(q,t,\Delta s,\ell) =
\begin{cases}
\beta_{\mathrm f}(t,\Delta s,\ell) & \text{if } m(q)=1,\\
\beta_{\mathrm b}(t,\ell) & \text{if } m(q)=0.
\end{cases}
\label{eq:sgam3}
\end{equation}

The attention keys at each layer consist of text tokens, condition tokens \((C_T(I_{\mathrm f}), S_T(I_{\mathrm b}))\), and latent tokens. We apply the resulting attention bias exclusively to the subset corresponding to \(S_T(I_{\mathrm b})\), whose key indices are denoted by \(\mathcal{K}_{\mathrm b}\):
\begin{equation}
B^{(\ell)}_{q,k}(t,\Delta s) =
\begin{cases}
\gamma(q,t,\Delta s,\ell) & \text{if } k \in \mathcal{K}_{\mathrm b},\\
0 & \text{otherwise}.
\end{cases}
\label{eq:sgam4}
\end{equation}

where \(B^{(\ell)}(t,\Delta s)\) denotes the resulting attention bias matrix. The final attention at layer \(\ell\) is computed as
\begin{equation}
\tilde{A}^{(\ell)} = \mathrm{softmax}\!\left(
\frac{Q^{(\ell)} K^{(\ell)\top}}{\sqrt{d_k}}
+ B^{(\ell)}(t,\Delta s)
\right),
\label{eq:sgam_attention}
\end{equation}
where \(Q^{(\ell)}\) and \(K^{(\ell)}\) denote the query and key matrices, respectively, and \(d_k\) is the key dimension. This formulation enables adaptive modulation of attention to the background style tokens according to the style gap, spatial region, and diffusion timestep, as illustrated in Fig.~\ref{fig:fig8}.

\subsection{Batch Construction for SSCGL}
\label{batch}

\noindent SSCGL operates on factor-specific relations induced within each mini-batch. We therefore construct each batch to ensure that both style-sharing and content-sharing relations are consistently available at every training step, rather than relying on incidental matches from random sampling. Specifically, each batch contains 32 style tuples and 32 content tuples, where each tuple is constructed to introduce complementary style and content relations for building $\mathcal{E}_S$ and $\mathcal{E}_C$.

\paragraph{Tuple construction.}
A style tuple starts from an anchor-positive pair that shares style but differs in content. We then add samples that share content with either the anchor or positive while differing in style, together with auxiliary samples that differ from the anchor in both factors. For example, given dog/Gogh and cat/Gogh as the anchor-positive pair, dog/Monet and cat/Picasso provide content-matched samples with different styles. A content tuple follows the same construction with the roles of style and content exchanged. This ensures that both style-sharing and content-sharing relations are explicitly introduced into each training batch.

\paragraph{Batch-wide graph construction.}
The tuples determine the composition of each batch, while the graph itself is constructed over the entire batch. After merging all tuples and removing duplicate samples, we connect every pair that satisfies the corresponding style or content relation, regardless of which tuple the samples originate from. This allows samples from different tuples, including auxiliary samples, to form additional edges and enables SSCGL to exploit all valid within-batch relations rather than only the pairs explicitly introduced during tuple construction.

In practice, each batch contains 1,655 unique images on average and yields 2,094 style edges and 1,591 content edges. Among these, 67\% of the style edges and 57\% of the content edges are formed across different tuples, demonstrating that batch-wide graph construction substantially expands the relational supervision available to SSCGL.

\begin{algorithm}[H]
\caption{Batch Construction for SSCGL}
\label{alg:sampling}
\begin{algorithmic}[1]
\Require Dataset $\mathcal{D}$ with style/content labels $(\phi_S,\phi_C)$; tuples per factor $T=32$; matched samples per seed $K_m=4$; auxiliary samples $K_a=16$
\Ensure Batch $\mathcal{B}$, relation matrices $\mathcal{E}_S,\mathcal{E}_C$, pairwise similarities $s,c$

\Function{Tuple}{$\mathcal{D}, A, B$}
    \State $I_a \sim \mathcal{U}(\mathcal{D})$
    \State $(\alpha,\beta) \gets \bigl(\phi_A(I_a),\,\phi_B(I_a)\bigr)$
    \State $I_p \sim \mathcal{U}\bigl(\{I\in\mathcal{D}\setminus\{I_a\} \mid \phi_A(I)=\alpha,\ \phi_B(I)\neq\beta\}\bigr)$
    \State $\beta' \gets \phi_B(I_p)$
    \State $\mathcal{M}_a \sim \mathcal{U}_{K_m}\bigl(\{I\in\mathcal{D} \mid \phi_A(I)\neq\alpha,\ \phi_B(I)=\beta\}\bigr)$
    \State $\mathcal{M}_p \sim \mathcal{U}_{K_m}\bigl(\{I\in\mathcal{D} \mid \phi_A(I)\neq\alpha,\ \phi_B(I)=\beta'\}\bigr)$
    \State $\mathcal{R} \sim \mathcal{U}_{K_a}\bigl(\{I\in\mathcal{D} \mid \phi_A(I)\neq\alpha,\ \phi_B(I)\notin\{\beta,\beta'\}\}\bigr)$
    \State \Return $\{I_a,I_p\}\cup\mathcal{M}_a\cup\mathcal{M}_p\cup\mathcal{R}$
\EndFunction

\Statex
\State $\mathcal{T}_S \gets \bigcup_{t=1}^{T}\Call{Tuple}{\mathcal{D},S,C}$
\State $\mathcal{T}_C \gets \bigcup_{t=1}^{T}\Call{Tuple}{\mathcal{D},C,S}$
\State $\mathcal{B} \gets \mathrm{Unique}\bigl(\mathcal{T}_S\cup\mathcal{T}_C\bigr)$

\Statex
\For{$i,j\in\mathcal{B}$}
    \State $\mathcal{E}_{S,ij} \gets \mathbbm{1}\!\left[i\neq j \,\wedge\, \phi_S(i)=\phi_S(j) \,\wedge\, \phi_C(i)\neq\phi_C(j)\right]$
    \State $\mathcal{E}_{C,ij} \gets \mathbbm{1}\!\left[i\neq j \,\wedge\, \phi_C(i)=\phi_C(j) \,\wedge\, \phi_S(i)\neq\phi_S(j)\right]$
\EndFor

\Statex
\State Compute $s,c\in\mathbb{R}^{|\mathcal{B}|\times|\mathcal{B}|}$ from frozen DINOv3 features
\State \Return $\mathcal{B},\mathcal{E}_S,\mathcal{E}_C,s,c$
\end{algorithmic}
\end{algorithm}

\section{Implementation Details}
\label{app:impl}

\subsection{Training Details}
Stage 1 uses a frozen DINOv3 ViT-L/16 backbone with two-layer MLP projection heads (Linear-ReLU-Linear) for style and content, where the input, hidden, and output dimensions are all 1024. Our proposed SSCGL is applied on top of the DINOv3 features through these style and content heads, which take features from layers [12,14] and [18,20], respectively, averaged across layers. Each batch is constructed from 32 style tuples and 32 content tuples so that both $E_S$ and $E_C$ are non-empty. The style similarity $s$ is computed from Gram statistics of DINOv3 layers [12,14], and the content similarity $c$ from patch-wise cosine similarity at layers [18,20]. Both are clipped to $[0,1]$, and self-pairs are excluded. Projected representations are constrained to the unit ball by norm clipping. We optimize with AdamW (learning rate $1\mathrm{e}{-3}$, weight decay $1\mathrm{e}{-4}$) for 40 epochs on a single H100 GPU, taking approximately 1 hour.

Stage 2 fine-tunes FLUX.1-dev~\citep{blackforestlabs2024flux} on ChameleonDataset$_{\mathrm{tr}}$ using bucketed resolutions (512, 768, 1024). We use LoRA (rank 16) with Prodigy (learning rate $1.0$) and jointly optimize lightweight modules including the projection layers and SGAM. The projection layers map 1024-dimensional features to 3072. Foreground and background DINO features are extracted from layers [18,20] and [12,14], respectively. SGAM consists of two three-layer MLP branches for foreground and background. The foreground branch takes the concatenation of a 256-dimensional sinusoidal time embedding and a 128-dimensional sinusoidal style-gap embedding, resulting in a \(384 \rightarrow 128 \rightarrow 128 \rightarrow 19\) architecture, whereas the background branch remains conditioned only on the 256-dimensional time embedding with a \(256 \rightarrow 128 \rightarrow 128 \rightarrow 19\) architecture. Both branches produce block-wise modulation logits for the 19 transformer blocks. Training uses bf16 mixed precision with a total batch size of 4 on 4$\times$H100 GPUs. We train for 40000 iterations. Total training takes approximately 22 hours.

\subsection{Compute and Efficiency Analysis}
\label{compute}
\begin{table}[H]
\centering
\footnotesize
\setlength{\tabcolsep}{7pt}
\renewcommand{\arraystretch}{1.15}
\caption{
\textbf{Trainable parameter budget.}
Percentages are relative to the frozen DiT backbone.
LoRA adapters account for most trainable parameters, while
SGAM introduces only a negligible overhead.
}
\label{tab:params}
\begin{tabular}{lrr}
\toprule
Module & Params & Backbone ratio \\
\midrule
LoRA adapters & $18.7$M & $0.16\%$ \\
SGAM & $0.1$M & $<0.01\%$ \\
Projection heads  & $6.3$M & $0.05\%$ \\
\midrule
Total trainable & $25.1$M & $0.21\%$ \\
\bottomrule
\end{tabular}
\end{table}

\begin{table}[H]
\centering
\footnotesize
\setlength{\tabcolsep}{3.5pt}
\renewcommand{\arraystretch}{1.15}
\caption{
\textbf{Inference efficiency.}
(a) We measure the overhead of dual-anchor DINO conditioning during DiT denoising.
Base denotes the standard DiT token sequence, consisting of image latent and text tokens,
whereas +DINO additionally appends the $392$ dual-anchor condition tokens of Fig.~\ref{fig:fig6},
composed of $196$ background style tokens $S_T(I_b)$ and $196$ foreground content tokens $C_T(I_f)$.
Adding these tokens introduces only a modest latency increase ($\Delta$).
(b) We further report the end-to-end runtime and peak memory of the full model.
All measurements use FP16 on a single H100 80GB with a $28$-step sampler.
}
\label{tab:inference_cost}

\textit{(a) DiT denoising overhead}

\vspace{2pt}
\begin{tabular}{l|cc|ccc|ccc}
\toprule
& \multicolumn{2}{c|}{Sequence length}
& \multicolumn{3}{c|}{Per-step latency}
& \multicolumn{3}{c}{28-step runtime} \\
Resolution
& Base & +DINO
& Base & +DINO & \textcolor{gray}{$\Delta$}
& Base & +DINO & \textcolor{gray}{$\Delta$} \\
\midrule
$512 \times 512$
& $1{,}280$ & $1{,}672$
& $0.281$\,s & $0.289$\,s & \textcolor{gray}{\scriptsize$+0.008$\,s ($+2.8\%$)}
& $7.86$\,s & $8.10$\,s & \textcolor{gray}{\scriptsize$+0.24$\,s ($+3.1\%$)} \\
$1024 \times 1024$
& $4{,}352$ & $4{,}744$
& $1.052$\,s & $1.099$\,s & \textcolor{gray}{\scriptsize$+0.047$\,s ($+4.5\%$)}
& $29.45$\,s & $30.77$\,s & \textcolor{gray}{\scriptsize$+1.32$\,s ($+4.5\%$)} \\
\bottomrule
\end{tabular}

\vspace{5pt}

\textit{(b) End-to-end inference cost}

\vspace{2pt}
\begin{tabular}{lcc}
\toprule
Resolution & Runtime & Peak memory \\
\midrule
$512 \times 512$   & $10.40$\,s & $36.2$\,GB \\
$768 \times 768$   & $19.83$\,s & $37.1$\,GB \\
$1024 \times 1024$ & $37.95$\,s & $38.5$\,GB \\
\bottomrule
\end{tabular}

\end{table}

\subsection{External Models and Licenses}
\label{license}
All external models and assets used in this work, including FLUX.1-dev~\citep{blackforestlabs2024flux}, Qwen~\citep{bai2025qwen3,wu2025qwen}, SAM3~\citep{carion2026sam}, and DINOv3~\citep{simeoni2025dinov3}, are publicly available and properly credited in the main paper. In particular, FLUX.1-dev is released under the FLUX.1 [dev] Non-Commercial License, and our use is restricted to non-commercial research purposes.

\subsection{User Study}
\label{user_study}
We conduct the user study with 50 participants who have AI-related research backgrounds. No monetary compensation is provided. The study includes 30 foreground-background cases selected from ChameleonDataset$_{\text{ev}}$, covering 30 distinct foreground objects and 27 background style families. Whenever possible, we select at most one case from each style family to avoid overrepresenting any particular style. The complete set of foreground-style combinations is provided in Tab.~\ref{tab:userstudy_assets}.

In each trial, participants view five anonymized results (A--E) side by side, together with the foreground object and the background image with the target region outlined, and select the single best result for each criterion. The ordering of the five anonymized results is randomized for each trial to reduce position bias. The study interface is shown in Fig.~\ref{fig:userstudy_ui}. The evaluation criteria follow the same definitions as the VLM rubric in Appendix~\ref{appendix:vlm-prompt}. The overall criterion evaluates whether the composition and harmonization are naturally integrated and visually plausible. Win rates for all five methods are reported in Tab.~\ref{tab:userstudy}.

\begin{figure}[t]
    \centering
    \includegraphics[width=\linewidth]{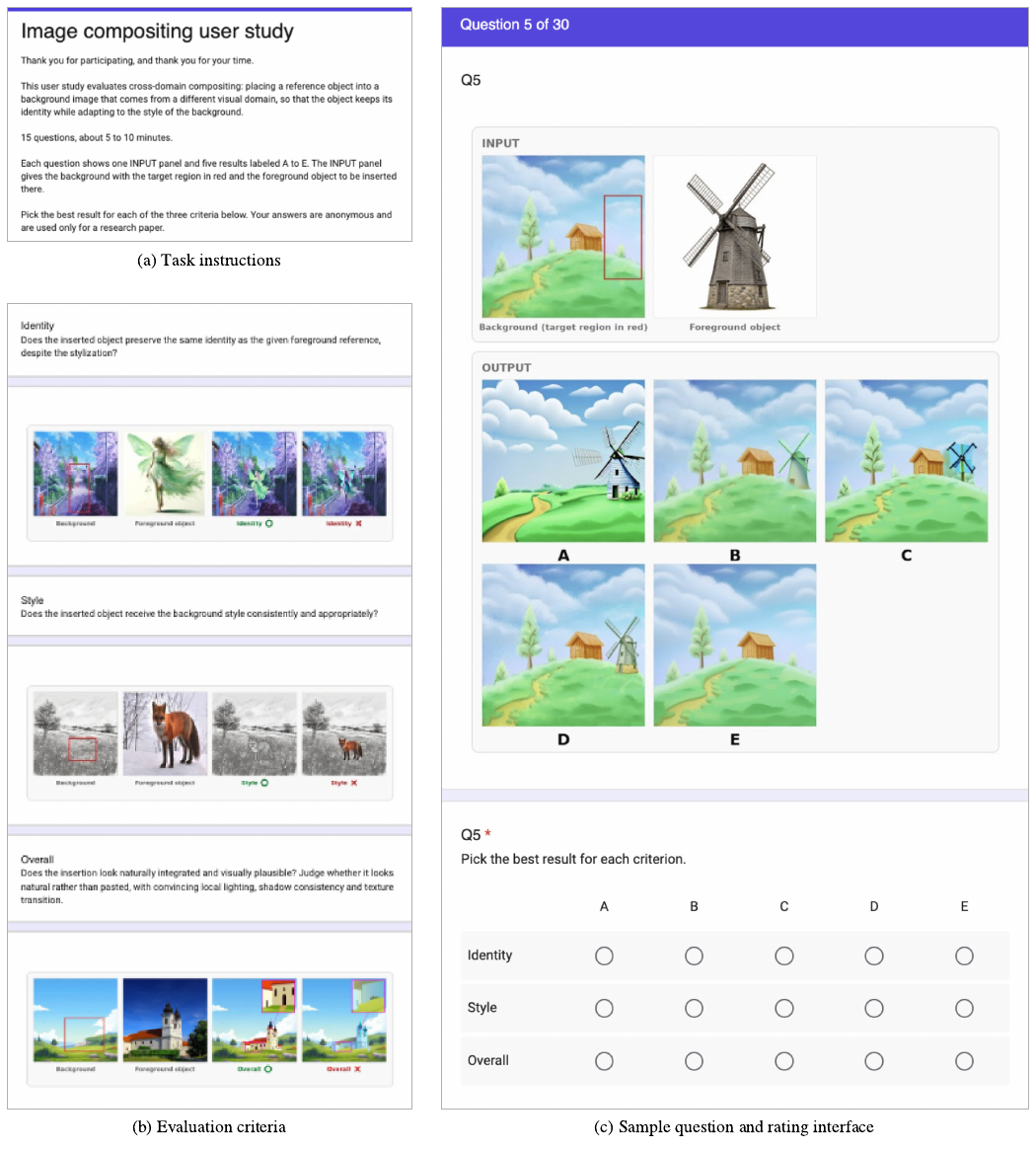}
    \caption{User study interface. Participants compare five anonymized results (A--E) and select the best result for each evaluation criterion.}
    \label{fig:userstudy_ui}
\end{figure}

\begin{table}[H]
    \centering
    \caption{Foreground objects and background style families used in the user study.}
    \label{tab:userstudy_assets}
    \setlength{\tabcolsep}{3pt}
    \renewcommand{\arraystretch}{1.05}
    \scriptsize
    \begin{tabular}{c l l @{\hspace{8pt}} c l l}
        \toprule
        \# & Foreground & Background style &
        \# & Foreground & Background style \\
        \midrule
        1  & Motorcycle        & Flat/Cartoon       & 16 & Woman in kimono   & Ukiyo-e \\
        2  & Whale             & Neon Line Art      & 17 & Pickup truck      & Pixel Art \\
        3  & Jellyfish         & Line Art           & 18 & Golden retriever  & Colored Pencil \\
        4  & Scooter           & Etching            & 19 & Popsicle          & 3D Render \\
        5  & Windmill          & Painterly          & 20 & Pumpkin           & Surrealism \\
        6  & Rabbit            & Anime/Cel          & 21 & Polar bear        & Watercolor \\
        7  & Cart              & Painterly          & 22 & Running courier   & Long Exposure \\
        8  & Sushi             & Flat/Cartoon       & 23 & Astronaut         & Astrophotography \\
        9  & Person            & Backlit Silhouette & 24 & Lobster           & Mosaic \\
        10 & Robot             & Pop Art            & 25 & Calligraphy text  & Floral Pattern \\
        11 & Butterfly         & Stained Glass      & 26 & Sports car        & Vaporwave \\
        12 & Mammoth           & Low-Poly           & 27 & Herd of deer      & Halftone \\
        13 & Easel with canvas & Art Nouveau        & 28 & Hot air balloon   & Oil Painting \\
        14 & Desk globe        & Flat/Cartoon       & 29 & Woman portrait    & Marbling \\
        15 & Dragonfly         & Monochrome Photo   & 30 & Koi fish          & Ink Splatter \\
        \bottomrule
    \end{tabular}
\end{table}

\clearpage

\section{Ablation Studies}
\label{app:abl}

\subsection{Encoder}
\label{app:abl:encoder}
In this section, we ablate various semantic encoders, including CLIP~\citep{radford2021learning}, SigLIP2~\citep{tschannen2025siglip}, and CSD~\citep{somepalli2024investigating}, to analyze the choice of DINOv3 in Tab.~\ref{tab:why_dinov3}. Although CSD is trained for understanding and extracting style descriptors from images, its representations still exhibit noticeable style-content entanglement. In contrast, the stronger semantic representations of DINOv3 provide a more favorable starting point for disentanglement prior to training the projection heads. 

We further analyze the effect of different DINOv3 layers in Tab.~\ref{tab:dinov3_layers}. The results justify our selection of separate layer ranges for the style and content heads. Larger margins between style and content similarity indicate representations that are easier to disentangle, which justifies our selection of different layer ranges for the style and content heads in ChameleonEncoder.

\begin{table}[H]
\centering
\footnotesize
\setlength{\tabcolsep}{5pt}
\renewcommand{\arraystretch}{1.15}
\caption{
\textbf{Why DINOv3?}
Disentanglement margin
$\Delta = \langle a,p\rangle - \langle a,hn\rangle$
on dual-anchor test splits (200 samples each), comparing frozen off-the-shelf encoders without any projection head.
Here, $a$ denotes the anchor feature, $p$ the positive feature, and $hn$ the hard-negative feature.
On the style-paired split, $p$ shares the same style as $a$, while $hn$ shares the same content but a different style.
On the content-paired split, the roles are reversed.
A larger $\Delta$ indicates better separation of the target attribute from the confounding one.
DINOv3 achieves the strongest overall disentanglement margin, particularly on the content-paired split, indicating that its representation provides the most suitable foundation for applying SSCGL.}
\label{tab:why_dinov3}
\setlength{\tabcolsep}{4pt}
\begin{tabular}{l|ccc|ccc}
\toprule
& \multicolumn{3}{c|}{Style-paired split}
& \multicolumn{3}{c}{Content-paired split} \\
\cmidrule(lr){2-4} \cmidrule(lr){5-7}
Encoder
& $\langle a,p\rangle \uparrow$
& $\langle a,hn\rangle \downarrow$
& $\Delta \uparrow$
& $\langle a,p\rangle \uparrow$
& $\langle a,hn\rangle \downarrow$
& $\Delta \uparrow$ \\
\midrule
CLIP ViT-L/14~\citep{radford2021learning}
& $0.848$ & $0.950$ & $-0.102$
& \textbf{0.949} & $0.844$ & $+0.105$ \\

SigLIP2 L/16-256~\citep{tschannen2025siglip}
& $0.755$ & $0.923$ & $-0.168$
& $0.927$ & $0.757$ & $+0.170$ \\

CSD ViT-L~\citep{somepalli2024investigating}
& $0.836$ & $0.906$ & $-0.071$
& $0.906$ & $0.838$ & $+0.068$ \\

\midrule
DINOv3~\citep{simeoni2025dinov3}
& \textbf{0.871} & \textbf{0.903} & \textbf{-0.032}
& $0.916$ & \textbf{0.660} & \textbf{+0.257} \\
\bottomrule
\end{tabular}
\end{table}

\begin{table}[H]
\centering
\footnotesize
\setlength{\tabcolsep}{5pt}
\renewcommand{\arraystretch}{1.15}
\caption{
\textbf{Layer selection for the style and content heads.}
Disentanglement margin
$\Delta = \langle a,p\rangle - \langle a,hn\rangle$
on dual-anchor test splits (200 samples each), comparing different layer groups from frozen DINOv3 ViT-L/16.
Here, $a$ denotes the anchor feature, $p$ the positive feature, and $hn$ the hard-negative feature.
On the style-paired split, $p$ shares the same style as $a$, while $hn$ shares the same content but a different style. On the content-paired split, the roles are reversed.
Since all layer groups produce negative margins on the style-paired split, we select the mid-level layers (12--14), which yield the smallest negative margin.
In contrast, late layers (18--20) provide substantially stronger content separation on the content-paired split.
Accordingly, we use layers 12--14 for the style head $E_s$ and layers 18--20 for the content head $E_c$ in SSCGL.}
\label{tab:dinov3_layers}
\begin{tabular}{l|ccc|ccc}
\toprule
& \multicolumn{3}{c|}{Style-paired split}
& \multicolumn{3}{c}{Content-paired split} \\
\cmidrule(lr){2-4} \cmidrule(lr){5-7}
DINOv3 layers
& $\langle a,p\rangle \uparrow$
& $\langle a,hn\rangle \downarrow$
& $\Delta \uparrow$
& $\langle a,p\rangle \uparrow$
& $\langle a,hn\rangle \downarrow$
& $\Delta \uparrow$ \\
\midrule
12--14 (mid, $\rightarrow E_s$)
& \textbf{0.874}
& \textbf{0.902}
& $\mathbf{-0.028}$
& $0.903$
& $0.879$
& $+0.024$ \\

18--20 (late, $\rightarrow E_c$)
& $0.655$
& $0.911$
& $-0.257$
& \textbf{0.919}
& \textbf{0.672}
& $\mathbf{+0.247}$ \\
\bottomrule
\end{tabular}
\end{table}


\begin{table}[H]
\centering
\footnotesize
\setlength{\tabcolsep}{4.5pt}
\renewcommand{\arraystretch}{1.15}
\caption{
\textbf{Ablation on DINOv3 layer combinations.}
We compare different style/content layer combinations for Stage 1 disentanglement and Stage 2 downstream compositing performance.}
\label{tab:layer_combination}
\begin{tabular}{l|cc|ccccc}
\toprule
& \multicolumn{2}{c|}{Disentanglement}
& \multicolumn{5}{c}{Compositing} \\
\cmidrule(lr){2-3} \cmidrule(lr){4-8}
Style / Content layers
& Style $\Delta \uparrow$
& Content $\Delta \uparrow$
& CLIP-I $\uparrow$
& CSD $\uparrow$
& CLIP-T $\uparrow$
& AES $\uparrow$
& LPIPS $\downarrow$ \\
\midrule

18--20 / 12--14 (swapped)
& 0.8851
& 0.9269
& 0.8398
& 0.4425
& 0.1985
& 6.6708
& 0.4921 \\

24 / 24 (no selection)
& 0.2438
& 0.6795
& 0.8104
& 0.3981
& 0.1987
& 6.6540
& 0.5407 \\

12--14 / 18--20 (Ours)
& \textbf{0.9272}
& \textbf{0.9481}
& \textbf{0.8425}
& \textbf{0.4629}
& \textbf{0.2033}
& \textbf{6.6946}
& \textbf{0.4651} \\

\bottomrule
\end{tabular}
\end{table}

\begin{table}[h]
\centering
\caption{
\textbf{Effect of factor-specific weighting in SSCGL.}
We report the disentanglement margin $\Delta = \langle a, p \rangle - \langle a, h_n \rangle$
(Tab.~\ref{tab:why_dinov3}) of the trained ChameleonEncoder heads on the dual-anchor test splits (400 samples each). Uniform replaces the factor-specific similarities with constant weights on the same signed graph. By modulating the strength of each pairwise constraint according to its factor similarity, factor-specific weighting improves the margin in both spaces.}
\label{tab:weighting}
\small
\begin{tabular}{lcc}
\toprule
Edge weighting & Style $\Delta \uparrow$ & Content $\Delta \uparrow$ \\
\midrule
Uniform (signed only)   & 0.8966          & 0.9358          \\
Factor-specific (Ours)  & \textbf{0.9272} & \textbf{0.9481} \\
\midrule
Gain                    & +0.0306         & +0.0123         \\
\bottomrule
\end{tabular}
\end{table}

\subsection{SGAM}
\label{app:abl:stag}
\begin{figure}[H]
    \centering
    \includegraphics[width=\linewidth]{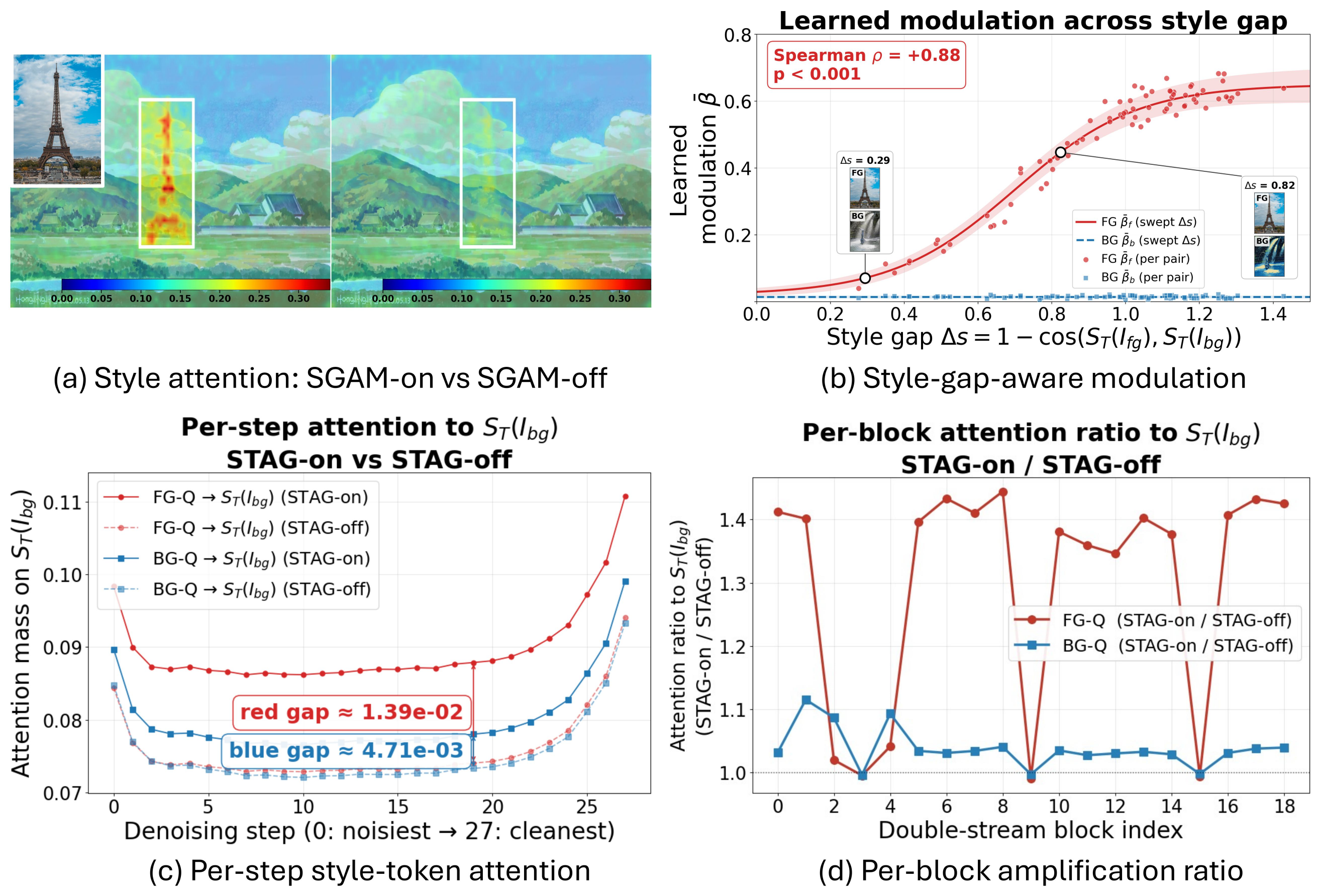}
    \vspace{-20pt}
    \caption{\textbf{Analysis of Style Gap-Aware Anisotropic Modulation (SGAM).} (a) Attention from the foreground query (white box) to $S_T(I_b)$ comparing SGAM-on and SGAM-off. SGAM concentrates style attention on the foreground. (b) Foreground modulation strength increases as the foreground--background style gap becomes larger. (c) SGAM learns timestep-adaptive style injection, with stronger stylization at later denoising steps. (d) Per-block amplification ratios show that SGAM strengthens and suppresses style injection block-by-block without collapse.}
    \label{fig:fig8}
    \vspace{-10pt}
\end{figure}

We ablate SGAM by comparing models with and without SGAM in Fig.~\ref{fig:fig8}. In (a), we visualize attention maps to show that SGAM effectively concentrates style attention on the foreground object region (Eiffel Tower). Although the spatial assignment is derived from a coarse binary bounding-box mask, the modulation remains spatially aligned with the object, achieving the intended foreground-focused style injection.

Moreover, (b) analyzes the style-gap-aware behavior of SGAM by measuring the learned foreground modulation across different foreground--background style gaps computed in the disentangled style space of ChameleonEncoder. We fix the foreground and vary the background from photorealistic images to diverse stylized domains, covering a broad range of foreground--background style gaps. The learned modulation shows a strong positive Spearman correlation with the style gap ($\rho=0.88$, $p<0.001$), becoming stronger as the foreground and background grow more stylistically different. This trend indicates that SGAM adaptively regulates foreground style injection according to the style gap.

In addition, (c) visualizes the timestep-adaptive behavior of SGAM across the diffusion process. We observe that stronger stylization emerges at later timesteps after the coarse structure is formed, consistent with the temporal tendency observed in diffusion-based style transfer~\citep{hu2024diffusest}. Rather than imposing this tendency through a fixed temporal schedule, SGAM learns timestep-dependent modulation tailored to cross-domain compositing, jointly balancing foreground identity preservation and stylization.

Finally, (d) shows the per-block modulation behavior. Rather than remaining uniformly active or inactive across all blocks, the modulation dynamically strengthens or suppresses style injection depending on the transformer block, indicating stable and non-collapsed modulation behavior. We also provide a qualitative component-wise ablation in Fig.~\ref{fig:fig_ablation}, where adding SGAM yields more seamless foreground--background integration.

We further examine the necessity of SGAM's adaptive style control by replacing each of its key adaptive components with a simpler alternative in Tab.~\ref{tab:stag_gating}. \textit{Masking} uses fixed spatial gates set to 1 and 0 for the foreground and background, respectively; \textit{Fixed schedule} replaces the learned temporal modulation with a predefined cosine schedule over diffusion timesteps; and \textit{Scalar gating} uses a single global gate shared across all spatial regions, timesteps, and transformer blocks. We also replace the learned style gap $\Delta s$ with a pixel-level statistic (\textit{Pixel-stat Gap}), computed as the distance between the channel-wise mean and standard deviation of $I_f$ and $I_b$ in LAB space following classical color transfer~\citep{reinhard2001color}, while keeping the SGAM architecture unchanged.

Masking and fixed scheduling improve style consistency but reduce identity preservation. Meanwhile, the scalar gate changes only marginally from 0.500 to 0.518 during training, indicating limited adaptation of style injection. Replacing the learned style gap with a simple pixel-level statistic also degrades performance, supporting the benefit of measuring the foreground--background style gap in the disentangled style space of ChameleonEncoder. In contrast, SGAM achieves the best performance across all metrics, demonstrating the effectiveness of adaptive style injection.

\begin{table}[H]
    \centering
    \caption{\textbf{Ablation of style modulation strategies on ChameleonDataset$_{\mathrm{ev}}$.} Comparison of SGAM with simpler modulation strategies and style-gap formulations under the same training and evaluation settings.}
    \label{tab:stag_gating}
    \small
    \begin{tabular}{lccccc}
        \toprule
        Method & CLIP-I $\uparrow$ & CSD $\uparrow$ & CLIP-T $\uparrow$ & AES $\uparrow$ & LPIPS $\downarrow$ \\
        \midrule
        Masking & 0.8258 & 0.4515 & 0.1945 & 6.5996 & 0.4896 \\
        Fixed schedule & 0.8285 & 0.4476 & 0.1942 & 6.5395 & 0.4845 \\
        Scalar gating & 0.8391 & 0.4394 & 0.1961 & 6.5563 & 0.4815 \\
        Pixel-stat Gap & 0.8210 & 0.4549 & 0.1922 & 6.6621 & 0.5391 \\
        \textbf{SGAM (Ours)} & \textbf{0.8425} & \textbf{0.4629} & \textbf{0.2033} & \textbf{6.6946} & \textbf{0.4651} \\
        \bottomrule
    \end{tabular}
\end{table}

\subsection{Qualitative Component-wise Ablation}
\label{app:qual_ablation}

\begin{figure}[H]
    \centering
    \includegraphics[width=0.85\linewidth]{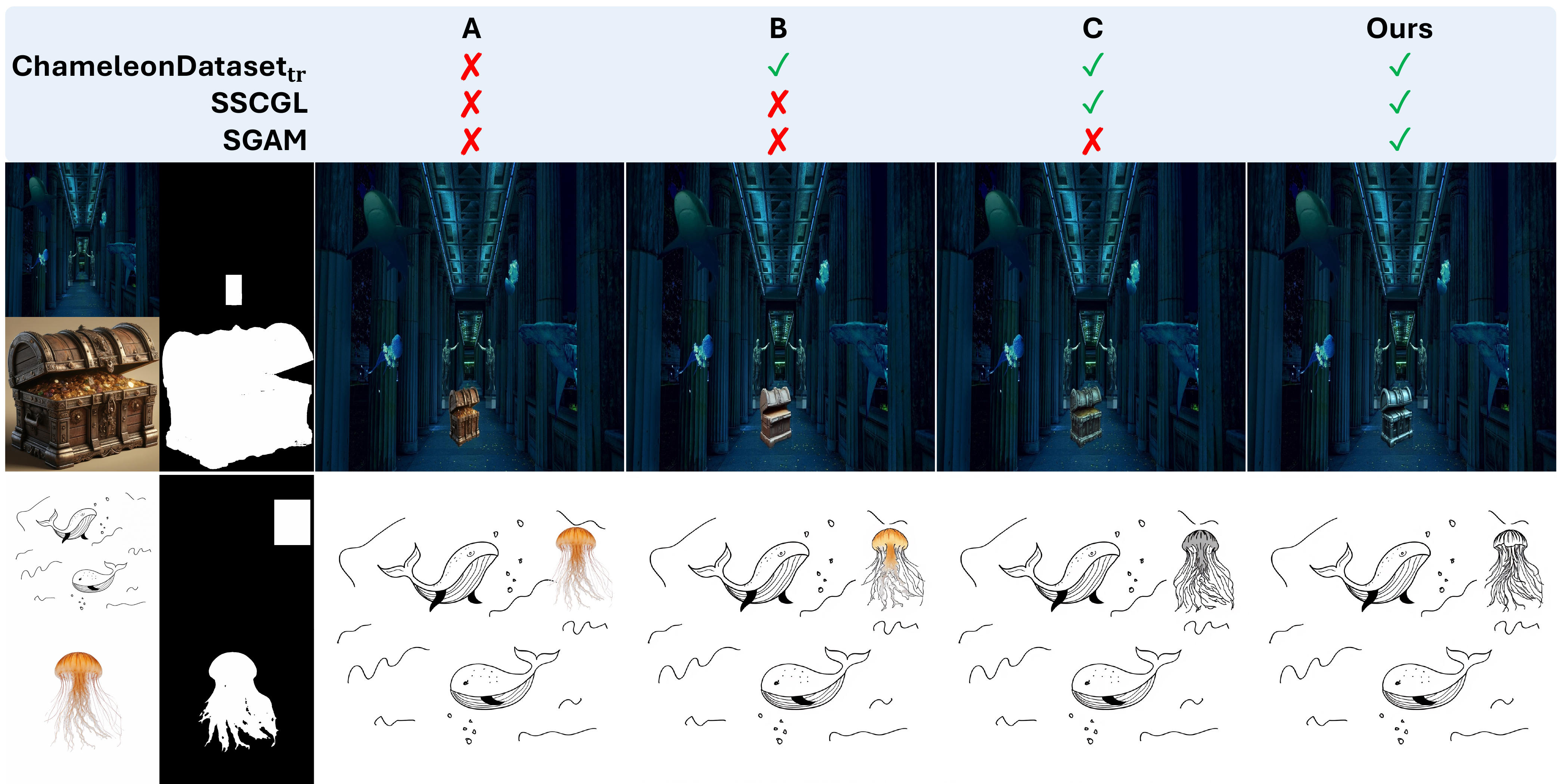}
    \caption{
    \textbf{Qualitative ablation of training data, SSCGL, and SGAM.}
    (A) Training on the in-domain DreamFuse data without our proposed modules learns object placement but largely ignores the surrounding context.
    (B) Training on ChameleonDataset$_{\mathrm{tr}}$ encourages adaptation to low-level background appearance cues, such as texture, but does not yet achieve consistent style matching.
    (C) SSCGL improves background-style consistency through disentangled style conditioning.
    Our full model uses SGAM to adaptively regulate style injection, yielding more coherent foreground-background integration.
    }
    \label{fig:fig_ablation}
\end{figure}

Finally, we provide a qualitative component-wise ablation in
Fig.~\ref{fig:fig_ablation}, progressively adding
ChameleonDataset$_{\mathrm{tr}}$, SSCGL, and SGAM.
The results visually complement the quantitative component-wise ablation
in Tab.~\ref{tab:ablation}.

\section{Inference Assumptions and Robustness}
\label{app:inference_robustness}

Our method takes a foreground image and a background image, while the target insertion region requires only coarse localization and can be specified using a user-drawn box. As shown in Tab.~\ref{tab:coarse_box}, the target localization mask can be replaced with a coarse user-drawn box with only minor performance changes.
This robustness follows from how spatial localization is used in our model.
Our mask-guided affine transformation uses the foreground and placement masks only to extract their axis-aligned bounding boxes.
Let the foreground box and target placement box be
$B_f=(x_f,y_f,w_f,h_f)$ and $B_t=(x_t,y_t,w_t,h_t)$,
where $(x,y)$ denotes the top-left coordinate and $(w,h)$ denotes the box size.
We compute the bbox-to-bbox affine transformation as

\begin{equation}
A =
\begin{bmatrix}
\frac{w_t}{w_f} & 0 & x_t-\frac{w_t}{w_f}x_f \\
0 & \frac{h_t}{h_f} & y_t-\frac{h_t}{h_f}y_f \\
0 & 0 & 1
\end{bmatrix}.
\end{equation}

For a foreground positional index $\mathbf{p}_f=[x,y,1]^\top$, the transformed index is given by $\mathbf{p}_f' = A\mathbf{p}_f$.
The transformation is applied only to the foreground latent positional indices, while the foreground latent values remain unchanged.
The masks $M_f$ and $M_{cp}$ are used only to obtain the bounding boxes and are not provided to the model as additional conditioning tokens or input channels.
Accordingly, the transformation depends on the coarse spatial extent of the regions rather than their exact mask boundaries.

\begin{table}[H]
    \centering
    \caption{\textbf{Robustness to coarse localization at inference.}
    Comparison between SAM 3 masks and coarse user-drawn boxes. IoU is measured against the SAM 3 masks (ref.) for $M_f$ and $M_{cp}$, respectively.}
    \label{tab:coarse_box}
    \vspace{3pt}
    \small
    \begin{tabular}{lccccc}
        \toprule
        Region at inference & IoU ($M_f$ / $M_{cp}$) & CLIP-I $\uparrow$ & CSD $\uparrow$ & LPIPS $\downarrow$ & Composition $\uparrow$ \\
        \midrule
        SAM 3 mask & ref. & 0.8425 & 0.4629 & 0.4651 & 2.11 \\
        User-drawn box & 0.798 / 0.702 & 0.8482 & 0.4508 & 0.4673 & 2.10 \\
        \bottomrule
    \end{tabular}
\end{table}

For foreground extraction, both training and inference use the same off-the-shelf SAM 3 model. The resulting crops can be incomplete or contain residual background regions due to segmentation errors. During training, the model is therefore exposed to such imperfect foreground crops while being supervised to reconstruct the complete foreground appearance in the composite. To further evaluate robustness to these imperfections, we deliberately perturb the foreground inputs using partial occlusion, dilation and erosion of $M_f$, and JPEG compression. Across these settings, Chameleon maintains stable identity preservation and style consistency, with CLIP-I and CSD decreasing by less than 0.02. This holds up to 25\% occlusion, 20-pixel mask dilation (IoU of 0.837 with the original SAM 3 mask), 30-pixel mask erosion (IoU of 0.648), and a JPEG compression ratio of $66\times$.

\section{Additional Quantitative Results}
\label{app:quant}

\subsection{Full Quantitative Results}
\label{app:quant_full}
Beyond the three primary metrics reported in the main paper, we additionally report the Aes score~\citep{schuhmann2022aesthetic}, which measures the overall aesthetic quality of the generated scene, and CLIP-T~\citep{hessel2021clipscore}, which evaluates alignment between the generated image and the text prompt, on both the AIComposer and TF-ICON benchmarks in Tab.~\ref{tab:comparison2} and Tab.~\ref{tab:tficon_bench}, respectively.
Moreover, we extend the quantitative comparison in Tab.~\ref{tab:foreground_metrics} from cross-domain compositing methods to specialized painterly image harmonization methods~\citep{cao2023painterly, lu2023painterly, niu2024painterly} and strong general-purpose image editing models.
While our method operates under the null-prompt setting, these editors are additionally given detailed text prompts describing the intended composite, as detailed below.
\textbf{Evaluation protocol for general-purpose editors.}
Each editor receives two images and a text prompt structured as follows:

{\small\ttfamily
IMAGE 1: a cut-out of the '\{obj\}' to insert, shown on a plain white background.\\
IMAGE 2: a background painting. A white rectangle marks where to place the object, with its normalized box coordinates given in the prompt.
}

The foreground is the same SAM-extracted cut-out used for all other methods, rendered on a plain white background.
The prompt then instructs the model to composite the object at the marked location and scale, harmonize its color, lighting, texture, and brush-stroke style to the background, remove the guide rectangle, and keep the rest of the background unchanged.
For GPT-Image-2~\citep{openai2026gptimage2} and Gemini-3-Pro-Image~\citep{google2025gemini3proimage}, the inputs are ordered as the foreground followed by the background.
For Qwen-Image-Edit, we reverse the order and adopt its picture-indexed phrasing, as we observed that it otherwise ignores the scene and generates the object in isolation.

\begin{table}[H]
\centering
\caption{\textbf{Quantitative comparison on the AIComposer benchmark.}}
\label{tab:comparison2}
\resizebox{\textwidth}{!}{%
\begin{tabular}{l|c|ccccc|cccc}
\toprule
Method & Res & LPIPS $\downarrow$ & CLIP-I $\uparrow$ & CSD $\uparrow$ & Aes $\uparrow$ & CLIP-T $\uparrow$ & Identity $\uparrow$ & Style $\uparrow$ & Composition $\uparrow$ & Avg\_total $\uparrow$ \\
\midrule
\multicolumn{11}{l}{\textit{In-domain compositing}} \\
\midrule
Dreamfuse~(ICCV'25)        & 1024 & 0.5290 & \underline{0.8603} & 0.3941 & \underline{6.9515} & \underline{0.2134} & 2.35 & 2.53 & 2.05 & 6.93 \\
SHINE~(ICLR'26)            & 1024 & 0.6513 & 0.8377 & 0.3512 & 6.9160 & 0.1999 & 2.05 & 2.40 & 1.99 & 6.44 \\
\midrule
\multicolumn{11}{l}{\textit{Two-stage compositing}} \\
\midrule
\makecell[l]{StyleSSP~(CVPR'25) \\ + Insert Anything~(AAAI'26)} & 1024 & 0.4682 & 0.8438 & \textbf{0.4898} & 6.7638 & 0.2083 & 2.26 & \textbf{2.83} & 1.97 & 7.07 \\
\midrule
\multicolumn{11}{l}{\textit{Cross-domain compositing}} \\
\midrule
TF-ICON~(ICCV'23)          & 512  & 0.6640 & 0.7493 & 0.3536 & 6.4443 & 0.1736 & 1.56 & 2.45 & 1.81 & 5.82 \\
TF-ICON~(ICCV'23)          & 768  & 0.6539 & 0.7700 & 0.3464 & 6.4217 & 0.1809 & 1.67 & 2.40 & 1.84 & 5.91 \\
TALE~(MM'24)               & 512  & 0.6164 & 0.7830 & 0.4681 & 6.9477 & 0.1893 & 1.72 & 2.72 & 2.00 & 6.44 \\
AIComposer~(ICCV'25)       & 1024 & \underline{0.4648} & 0.8575 & 0.4848 & 6.8638 & 0.2006 & \underline{2.42} & 2.77 & \underline{2.10} & \underline{7.30} \\
\midrule
\textbf{Ours}              & 1024 & \textbf{0.4556} & \textbf{0.8637} & \underline{0.4886} & \textbf{7.0296} & \textbf{0.2188} & \textbf{2.60} & \underline{2.81} & \textbf{2.23} & \textbf{7.64} \\
\bottomrule
\end{tabular}%
}
\end{table}

\begin{table}[H]
\centering
\caption{\textbf{Quantitative comparison on the TF-ICON benchmark.}}
\label{tab:tficon_bench}
\resizebox{\textwidth}{!}{%
\begin{tabular}{l|c|ccccc|cccc}
\toprule
Method & Res & LPIPS $\downarrow$ & CLIP-I $\uparrow$ & CSD $\uparrow$ & Aes $\uparrow$ & CLIP-T $\uparrow$ & Identity $\uparrow$ & Style $\uparrow$ & Composition $\uparrow$ & Avg\_total $\uparrow$ \\
\midrule
\multicolumn{11}{l}{\textit{In-domain compositing}} \\
\midrule
BLD~(SIGGRAPH'23)          & 512  & 0.7540 & 0.7331 & 0.4619 & 6.6360 & 0.2959 & 2.04 & 2.83 & 2.11 & 6.98 \\
Paint by Example~(CVPR'23) & 512  & 0.7501 & 0.7658 & 0.3138 & 6.7032 & 0.2969 & 2.17 & 2.46 & 2.18 & 6.81 \\
Insert Anything~(AAAI'26)  & 1024 & 0.7281 & 0.7719 & 0.3821 & 6.7440 & \textbf{0.3007} & 2.31 & 2.59 & 2.24 & 7.14 \\
SHINE~(ICLR'26)            & 512  & 0.7654 & 0.7155 & 0.4362 & 6.7360 & 0.2889 & 1.93 & 2.74 & 2.16 & 6.83 \\
SHINE~(ICLR'26)            & 1024 & 0.7737 & 0.7165 & 0.4627 & 6.7540 & 0.2891 & 2.02 & 2.71 & 2.22 & 6.95 \\
\midrule
\multicolumn{11}{l}{\textit{Cross-domain compositing}} \\
\midrule
TF-ICON~(ICCV'23)          & 512  & 0.7740 & 0.7551 & 0.4103 & 6.7080 & 0.2895 & 2.20 & 2.77 & 2.13 & 7.10 \\
TF-ICON~(ICCV'23)          & 768  & 0.7440 & 0.7581 & 0.4073 & 6.7240 & 0.2902 & 2.22 & 2.71 & 2.19 & 7.12 \\
TALE~(MM'24)               & 512  & 0.7990 & 0.5395 & \underline{0.4982} & \underline{6.8194} & 0.2422 & 2.09 & 2.96 & 2.27 & 7.33 \\
PrimeComposer~(MM'24)      & 512  & 0.7762 & 0.7510 & 0.4285 & 6.3930 & 0.2851 & 2.12 & 2.74 & 1.93 & 6.79 \\
AIComposer~(ICCV'25)       & 512  & 0.5025 & 0.7946 & 0.4674 & 6.6661 & 0.2853 & 2.19 & 2.83 & 1.98 & 7.00 \\
AIComposer~(ICCV'25)       & 1024 & \underline{0.4966} & \underline{0.8264} & 0.4826 & 6.8174 & 0.2908 & \underline{2.37} & 2.91 & 2.07 & 7.36 \\
\midrule
\textbf{Ours}              & 512  & 0.5018 & 0.8001 & 0.4728 & 6.7959 & 0.2872 & 2.34 & \textbf{2.99} & \underline{2.28} & \underline{7.61} \\
\textbf{Ours}              & 1024 & \textbf{0.4855} & \textbf{0.8296} & \textbf{0.4994} & \textbf{6.8340} & \underline{0.2983} & \textbf{2.46} & \underline{2.98} & \textbf{2.38} & \textbf{7.82} \\
\bottomrule
\end{tabular}%
}
\vspace{-10pt}
\end{table}

\begin{table}[H]
    \centering
    \caption{
    \textbf{Quantitative comparison on ChameleonDataset$_{\mathrm{ev}}$.}
    Beyond cross-domain compositing methods, we extend the comparison to specialized painterly harmonization methods and strong general-purpose image editing models, all under identical foreground, background, and mask inputs. Ours outperforms prior methods on most metrics, achieving identity preservation and stylization simultaneously (CLIP-I and CSD).}
    \label{tab:foreground_metrics}
    \small
    \setlength{\tabcolsep}{5pt}
    \renewcommand{\arraystretch}{1.05}
    \begin{tabular*}{\linewidth}{@{\extracolsep{\fill}}lccccc@{}}
        \toprule
        Method & LPIPS $\downarrow$ & CLIP-I $\uparrow$ & CSD $\uparrow$ & Aes $\uparrow$ & CLIP-T $\uparrow$ \\
        \midrule
        \multicolumn{6}{l}{\textit{Cross-domain compositing}} \\
        \midrule
        TF-ICON (ICCV'23)        & 0.6499 & 0.7407 & 0.3833 & 6.4497 & 0.1760 \\
        TALE (MM'24)             & 0.6415 & 0.7214 & \underline{0.4586} & 6.5296 & 0.1672 \\
        PrimeComposer (MM'24)    & 0.7292 & 0.7736 & 0.3745 & 6.1044 & 0.1954 \\
        AIComposer (ICCV'25)     & 0.4688 & 0.8275 & 0.4542 & 6.5855 & 0.1984 \\
        \midrule
        \multicolumn{6}{l}{\textit{Painterly image harmonization}} \\
        \midrule
        PHDNet (AAAI'23)            & 0.4877 & 0.7835 & 0.3913 & 6.1978 & 0.1871 \\
        PHDiffusion (MM'23) w/ Res  & 0.4733 & 0.7511 & 0.4123 & 6.4691 & 0.1876 \\
        PHDiffusion (MM'23) w/o Res & 0.4794 & 0.7371 & 0.4230 & 6.5030 & 0.1834 \\
        ArtoPIH (AAAI'24)           & \textbf{0.4548} & 0.7768 & 0.3480 & 6.2815 & 0.1947 \\
        \midrule
        \multicolumn{6}{l}{\textit{General-purpose image editing}} \\
        \midrule
        GPT-Image-2              & 0.6247 & \textbf{0.8578} & 0.4356 & 6.6722 & 0.1992 \\
        Gemini-3-Pro-Image       & 0.6584 & 0.8002 & 0.4470 & \underline{6.6824} & 0.2015 \\
        Qwen-Image-Edit          & 0.6940 & 0.7838 & 0.2874 & 6.6648 & \underline{0.2018} \\
        \midrule
        \textbf{Ours}            & \underline{0.4651} & \underline{0.8425} & \textbf{0.4629} & \textbf{6.6946} & \textbf{0.2033} \\
        \bottomrule
    \end{tabular*}
\end{table}

\clearpage

\subsection{Pose Adaptation Evaluation}
\label{app:pose_adaptation}

We evaluate pose and viewpoint adaptation under substantial viewpoint changes using a separately curated 50-image evaluation set constructed with our reverse pipeline. The source images do not overlap with those used in ChameleonDataset$_{\mathrm{tr}}$ or ChameleonDataset$_{\mathrm{ev}}$. Concretely, for each sample, we perturb the camera parameters of the foreground reference $I'_f$ to obtain $I''_f$ with a different pose and viewpoint, while using the original object in the real composite $I_c$ as the ground-truth target. For each resulting pair, we quantify the viewpoint change as the geodesic angular distance between the viewing directions of $I''_f$ and the target object in $I_c$, jointly accounting for azimuth and elevation. Across the 50 samples, the median of these angular distances is 79.7$^\circ$, with the middle 50\% ranging from 45.0$^\circ$ to 90.0$^\circ$, indicating substantial viewpoint variation. Camera distance is additionally scaled by $\times0.6$ in 12\% of the samples and by $\times1.8$ in 22\%. We evaluate foreground--scene compatibility using FOSScore~\citep{zhang2023foreground,chen2026carecom}, where a higher score indicates better compatibility between the generated foreground and the target scene. We additionally report CLIP-I between the generated object and the target object in $I_c$ to measure identity preservation under viewpoint changes.

\begin{table}[t]
    \centering
    \caption{
    \textbf{Quantitative evaluation of pose adaptation under substantial viewpoint changes.}
    FOSScore measures foreground--scene compatibility, while CLIP-I measures
    foreground identity preservation.
    w/ and w/o App. Var. denote training with and without data generated by the
    appearance-variation stage, respectively.
    }
    \label{tab:pose_adaptation}
    \resizebox{\linewidth}{!}{
    \begin{tabular}{lcccccccc}
        \toprule
        &
        \multicolumn{1}{c}{Baseline} &
        \multicolumn{5}{c}{Existing Methods} &
        \multicolumn{2}{c}{Ours} \\
        \cmidrule(lr){2-2}
        \cmidrule(lr){3-7}
        \cmidrule(lr){8-9}
        Metric &
        Copy \& Paste &
        PHDNet &
        TALE &
        ArtoPIH &
        PHDiffusion &
        AIComposer &
        w/o App. Var. &
        w/ App. Var. \\
        \midrule
        FOSScore $\uparrow$ &
        0.7427 &
        0.7475 &
        0.7441 &
        \underline{0.7740} &
        0.7281 &
        0.7667 &
        0.7522 &
        \textbf{0.8072} \\
        CLIP-I $\uparrow$ &
        \textbf{0.8360} &
        0.7420 &
        0.6990 &
        0.7230 &
        0.7331 &
        0.8023 &
        \underline{0.8306} &
        0.8155 \\
        \bottomrule
    \end{tabular}
    }
\end{table}

As shown in Tab.~\ref{tab:pose_adaptation}, Chameleon achieves the highest
FOSScore of 0.8072, indicating more effective adaptation of the perturbed
foreground to the target scene under substantial viewpoint changes.
Compared with the model trained without appearance variation, FOSScore
substantially increases from 0.7522 to 0.8072.
Despite this stronger adaptation, Chameleon maintains the highest CLIP-I
among the competing generative methods at 0.8155, measured between the
generated object and the target object in $I_c$.
These results show that appearance variation improves pose and viewpoint
adaptation while maintaining strong foreground identity.
A representative example is shown in Fig.~\ref{fig:fig13}(a).

\section{Additional Qualitative Results}
\label{app:qual}

\subsection{Extended Benchmark Comparisons}
\begin{figure}[t]
    \centering
    \includegraphics[width=\linewidth]{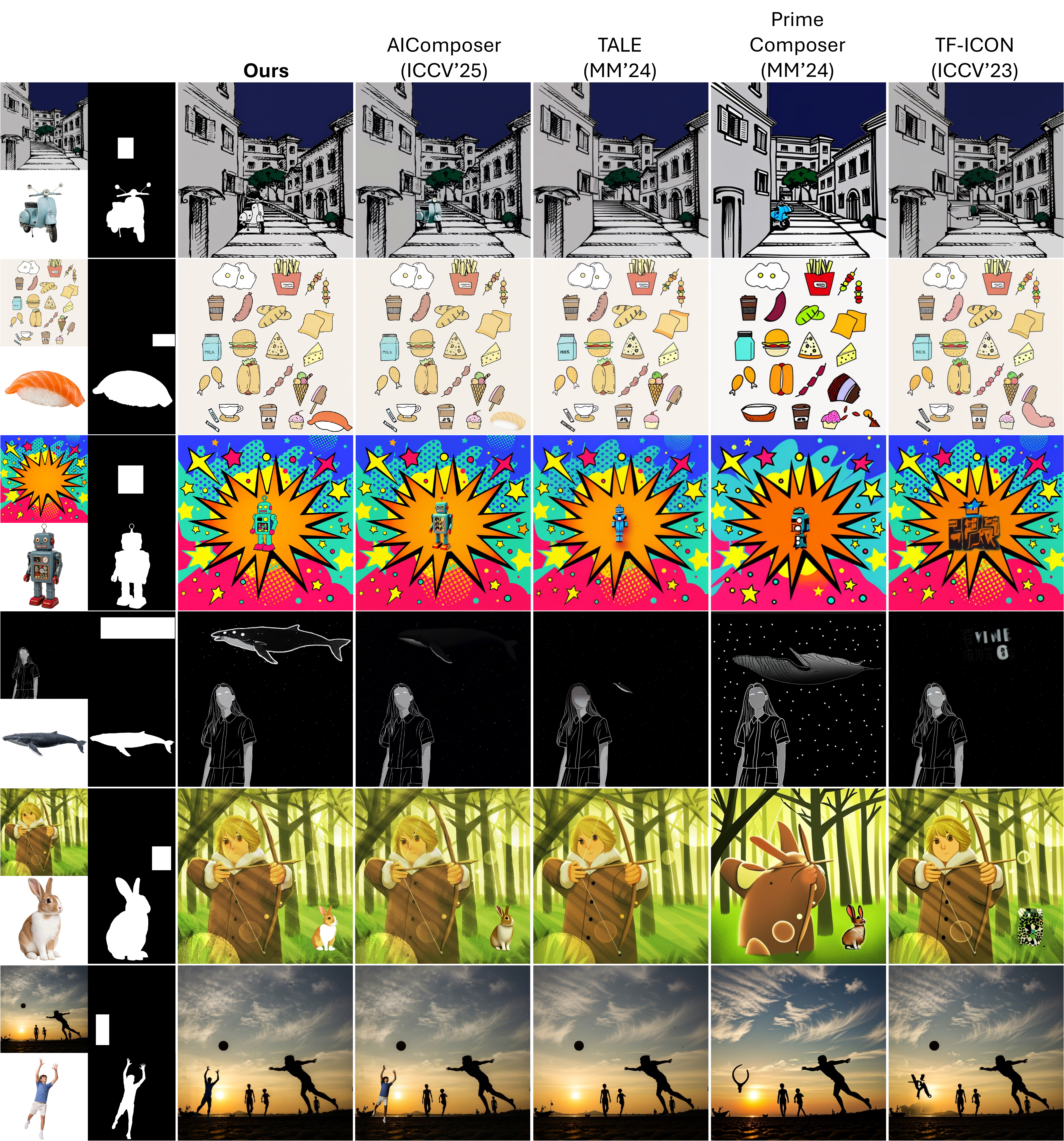}
    \vspace{-15pt}
    \caption{\textbf{Qualitative comparison with cross-domain baselines on our ChameleonDataset$_{\mathrm{ev}}$.} Unlike prior benchmarks that reuse backgrounds and foregrounds from existing datasets and pair them without regard to compositional plausibility, which often yields unrealistic scenarios, our benchmark designs the inserted object and its placement for each scene, enabling a more fine-grained and accurate evaluation. Across these diverse scenarios, our method harmonizes arbitrary styles and objects faithfully, demonstrating that it is general and robust.}
    \label{fig:fig_add1}
\end{figure}

\begin{figure}[t]
    \centering
    \includegraphics[width=\linewidth]{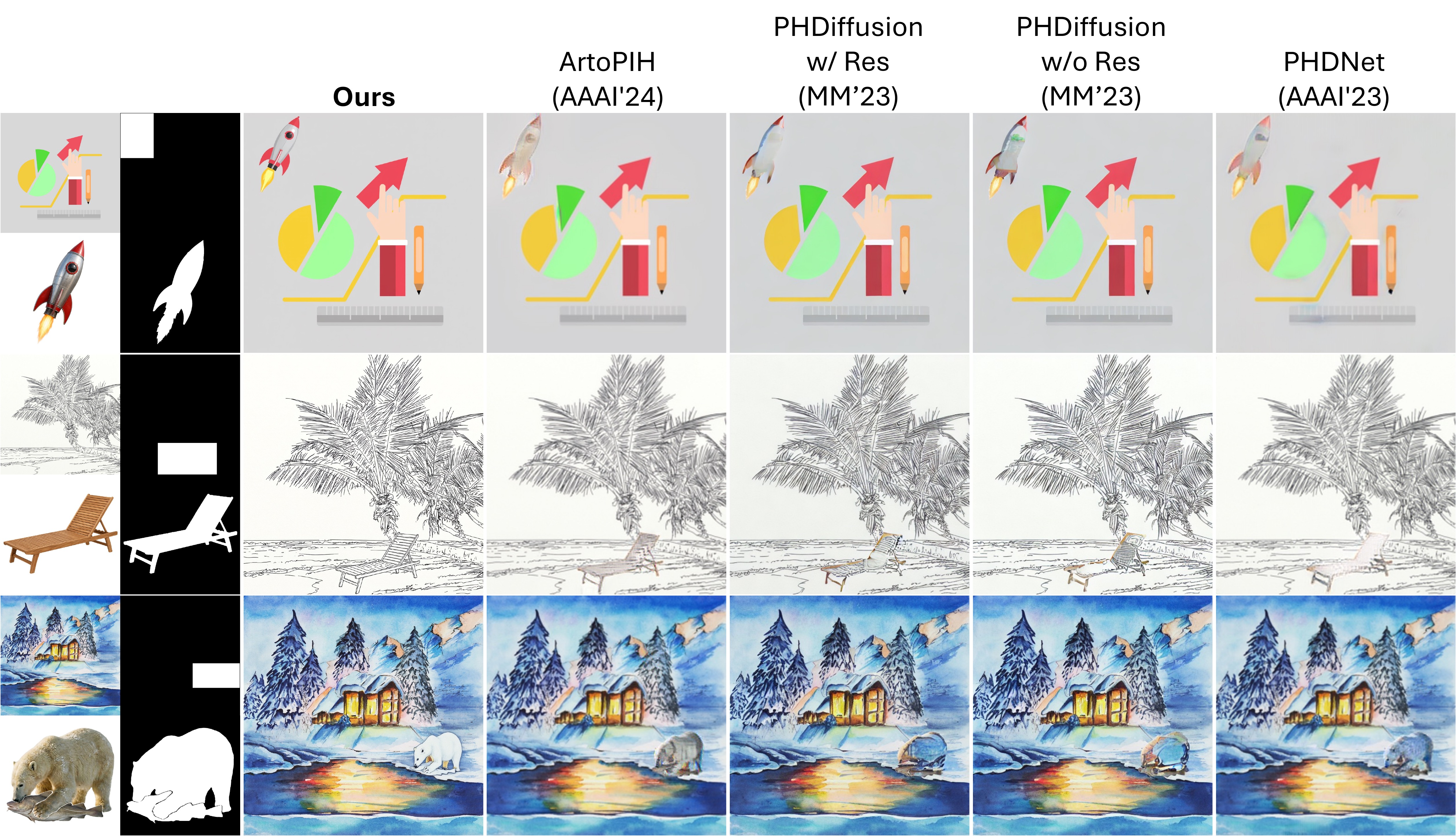}
    \vspace{-15pt}
    \caption{\textbf{Qualitative comparison with specialized painterly image harmonization methods on ChameleonDataset$_{\mathrm{ev}}$.} Painterly image harmonization methods start from a copy-and-pasted foreground with fixed placement and only align color and texture within the pasted region. Consequently, the foreground is merely blended into the background: it adopts the tone of the target domain but not its structural characteristics, and object identity is compromised under this blending. In contrast, our Chameleon preserves object identity while adapting the foreground to the structural style of the background, and further performs scene-aware pose adaptation, naturally seating the man on the bench (last row), which is unattainable under the copy-and-paste formulation of painterly harmonization.}
    \label{fig:fig_har}
\end{figure}

\begin{figure}[t]
    \centering
    \includegraphics[width=\linewidth]{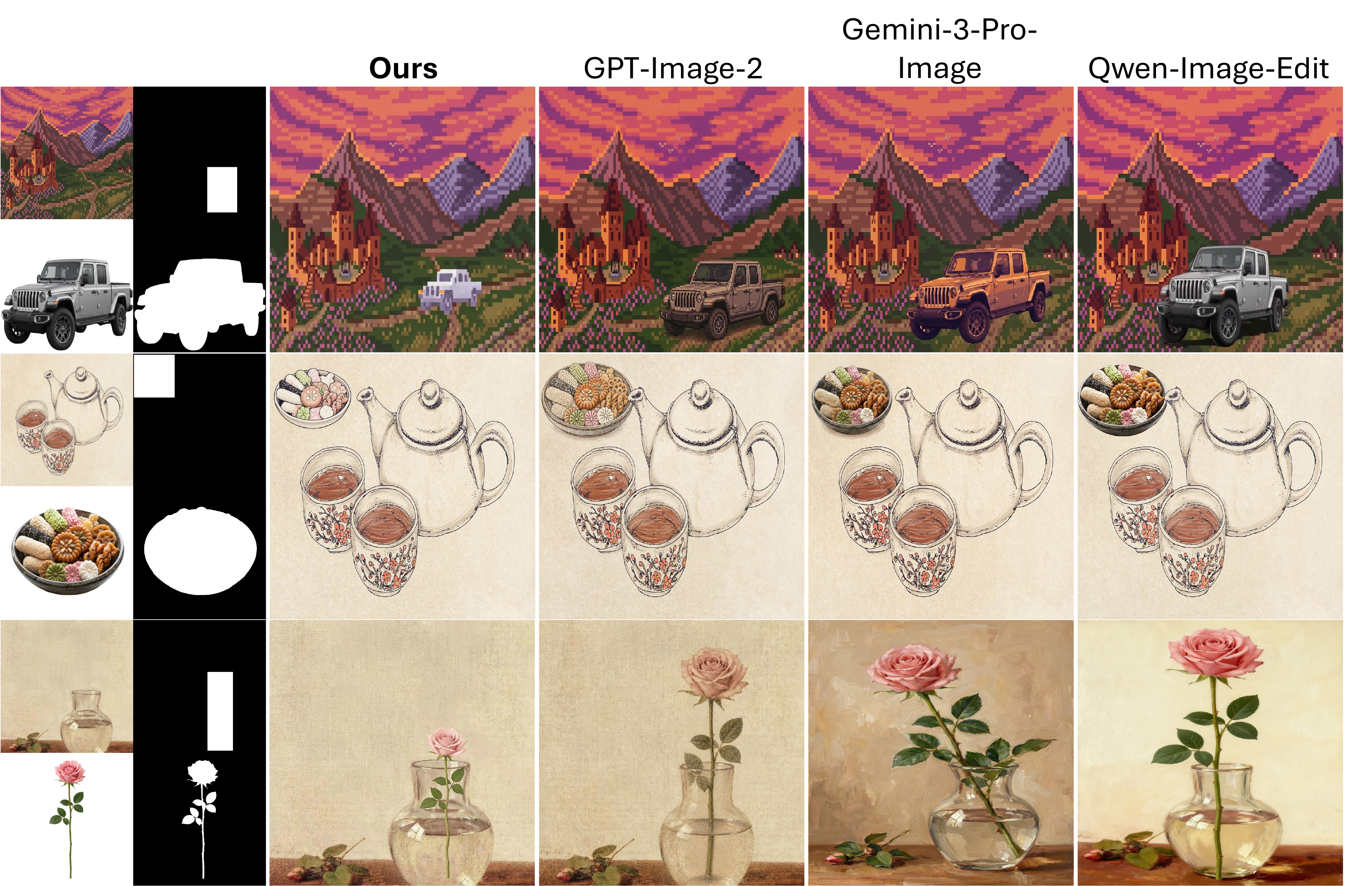}
    \vspace{-15pt}
    \caption{\textbf{Qualitative comparison with general-purpose image editing models on ChameleonDataset$_{\mathrm{ev}}$.} Although an identical mask specifying the insertion region is given, the general-purpose editing models disregard the user-specified size and location of the foreground. Their stylization is also incomplete, as the foreground fails to adopt the structural characteristics of unique domains such as pixel art (2nd row) or remains photorealistic without being fully stylized (3rd row). In contrast, our Chameleon faithfully follows the user-provided mask and composites the foreground in a style consistent with the background domain.}
    \label{fig:fig_commercial}
\end{figure}

We provide additional qualitative comparisons on ChameleonDataset$_{\mathrm{ev}}$ in Figs.~\ref{fig:fig_add1} , \ref{fig:fig_har}, and \ref{fig:fig_commercial}, which include diverse foreground objects and challenging styles such as pixel-art scenes. 
Our Chameleon consistently adapts foreground objects to the target background style while preserving natural appearance and compositional plausibility.

\begin{figure}[t]
    \centering
    \includegraphics[width=\linewidth]{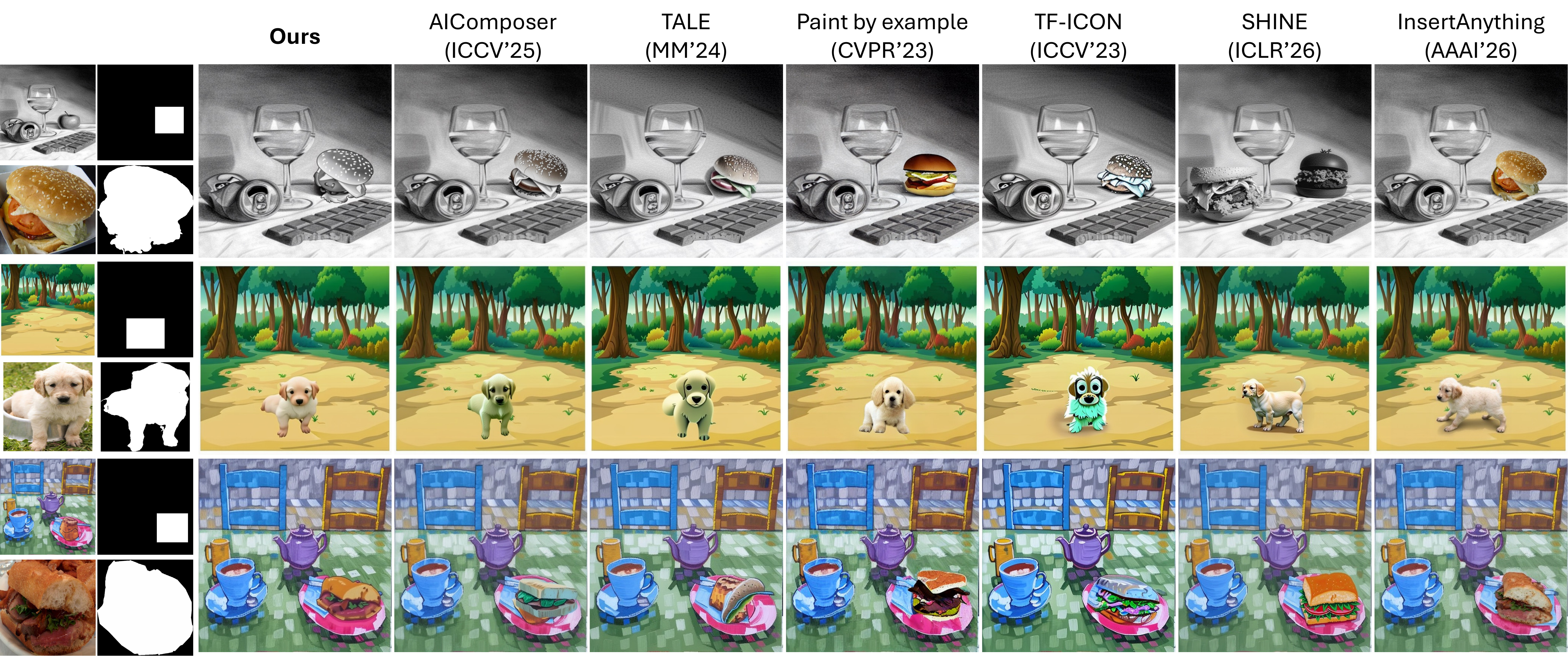}
    \vspace{-15pt}
    \caption{\textbf{Qualitative comparison with baselines on the TF-ICON benchmark.} The benchmark consists of three cross-domain categories: real-to-sketch, real-to-cartoon, and real-to-painting. We visualize representative scenes from each domain. Our method consistently achieves seamless integration and plausible compositing across diverse background domains and styles.}
    \label{fig:fig12}
\end{figure}

We also provide additional qualitative results on the TF-ICON benchmark in Fig.~\ref{fig:fig12} across all three domains, including sketch, cartoon, and painting. Across diverse domains, our method demonstrates robust cross-domain adaptation and seamless integration with the background. 

\FloatBarrier
\subsection{Scene-Aware Pose Adaptation and Generalized Cross-Domain Compositing}
\label{app:qual:challenging}

\begin{figure}[H]
    \centering
    \includegraphics[width=\linewidth]{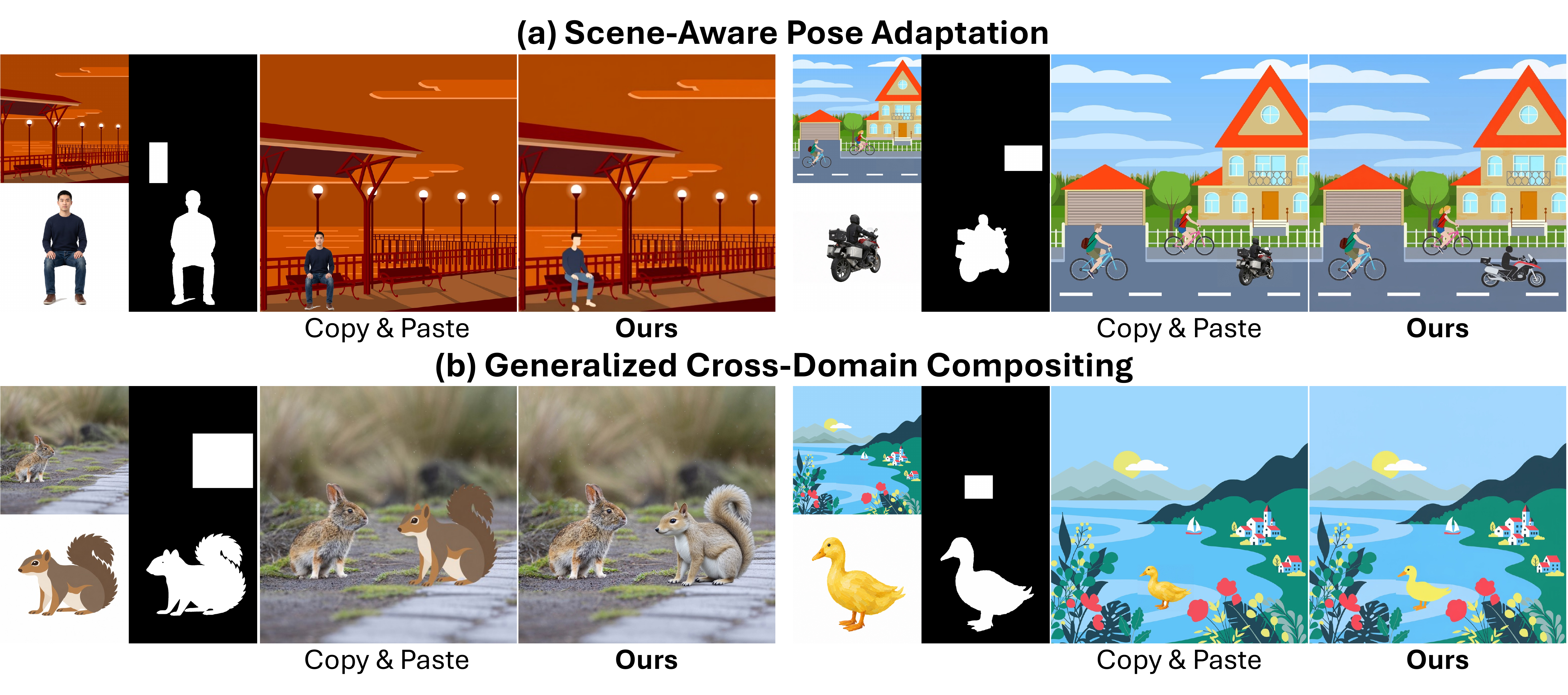}
    \vspace{-15pt}
    \caption{
    \textbf{Scene-aware pose adaptation and generalized cross-domain compositing.}
    (a) Chameleon exhibits scene-aware pose adaptation, adjusting the foreground pose
    to the surrounding scene context rather than directly copying and pasting the input
    foreground.
    (b) Chameleon further demonstrates generalization beyond the conventional
    photorealistic-to-stylized setting, including both stylized-to-photorealistic and
    stylized-to-stylized compositing.
    }
    \label{fig:fig13}
\end{figure}

Our reverse data construction pipeline (Fig.~\ref{fig:fig2}) allows controlled perturbations of the foreground while preserving the corresponding real-image target. In particular, the appearance variation step in Sec.~\ref{sec:dataset} perturbs factors such as azimuth, which provides training examples that encourage scene-aware pose adaptation. As illustrated in Fig.~\ref{fig:fig13}(a), Chameleon can adjust the foreground pose to better fit the surrounding scene context, rather than merely copying and pasting the input foreground. In addition, as described in Sec.~\ref{sec:stage_generation}, Chameleon extracts background style cues from DINO features and adaptively injects them into the foreground via SGAM, which appears to support generalization across different background domains. Consistent with this, Fig.~\ref{fig:fig13}(b) provides qualitative evidence that Chameleon extends beyond the conventional photorealistic-to-stylized setting, and can also handle stylized-to-photorealistic and stylized-to-stylized compositing. These results further suggest that Chameleon generalizes to unseen source and target domains.

\clearpage 
\section{VLM-based Protocols}
\label{appendix:vlm-prompt}

\subsection{VLM-based Evaluation Metric}

\paragraph{System / Role.}
\begin{quote}
You are a meticulous cross-domain image compositing evaluation assistant.

Your task is to evaluate a model output for cross-domain compositing using criterion-specific image priorities.
\end{quote}

\paragraph{Provided Inputs.}
The evaluator is given the following images:
\begin{enumerate}
    \item \textbf{f\_seg full image}: a clean object-centric reference on a white background.
    \item \textbf{background\_image full image}: the target scene whose overall style should guide stylization.
    \item \textbf{model\_output full image}: the final generated/composited result.
    \item \textbf{model\_output\_crop}: the cropped region inside the provided bounding box from the model output.
    \item \textbf{background\_mask image}: the source used to define the target region. Its white region is resized to the model output resolution before extracting the bounding box.
\end{enumerate}

\paragraph{Criterion-specific Image Priority.}
\begin{itemize}
    \item For \textbf{Identity Preservation}, focus primarily on the f\_seg full image and the model\_output\_crop.
    \item For \textbf{Style Transfer Consistency}, focus primarily on the background\_image full image and the model\_output\_crop.
    \item For \textbf{Composition and Harmonization}, focus primarily on the model\_output\_crop, and use the model\_output full image as secondary context.
\end{itemize}

\paragraph{Evaluation Rules.}
\begin{itemize}
    \item Use the stated priority images for each criterion.
    \item Do not require the same image pair for every criterion.
    \item For Identity Preservation, judge whether the cropped output object preserves the same object identity as the clean f\_seg reference.
    \item For Style Transfer Consistency, judge whether the cropped output object reflects the style of the full background scene appropriately.
    \item For Composition and Harmonization, judge whether the cropped output region looks naturally integrated and visually plausible, while using the full output only as secondary global context.
    \item Do not penalize based on regions unrelated to the provided bounding box.
    \item Do not let style transfer alone override identity preservation.
    \item Do not let identity similarity alone override poor harmonization.
\end{itemize}

\paragraph{Identity Preservation.}
\textbf{Question.}
Does the object inside the provided bounding box in the model output preserve the same identity as the object in the f\_seg reference image, despite the stylization? Consider whether the object remains recognizably the same in terms of its essential structure, shape, distinctive parts, and overall visual identity.

\textbf{Scoring Guide (0--3).}
\begin{itemize}
    \item \textbf{3}: The object clearly preserves the same identity as the f\_seg reference. Its structure, distinctive parts, and overall appearance remain highly recognizable despite stylization.
    \item \textbf{2}: The object mostly preserves the same identity, but there are minor ambiguities or slight structural/detail changes.
    \item \textbf{1}: The object weakly preserves the identity. Major parts or structure are altered, making it difficult to confidently match.
    \item \textbf{0}: The object does not preserve the identity. It appears as a different object or is unrecognizable.
\end{itemize}

\paragraph{Style Transfer Consistency.}

\textbf{Question.}
Does the object inside the provided bounding box in the model output receive the background style consistently and appropriately? Consider whether the stylization reflects the full background style well, not just blending background color, while avoiding excessive style leakage that unnaturally distorts the object or weakens its visual coherence.

\textbf{Scoring Guide (0--3).}
\begin{itemize}
    \item \textbf{3}: The object is stylized in a way that strongly and consistently reflects the background style. The style transfer is clear and appropriate, and there is no noticeable excessive style leakage or unnatural distortion.
    \item \textbf{2}: The object mostly reflects the background style, but the style transfer is slightly incomplete, uneven, or accompanied by minor style leakage that slightly affects coherence.
    \item \textbf{1}: The object shows weak or inconsistent style transfer from the background. The stylization is limited, patchy, or noticeably affected by style leakage that harms the object's coherence.
    \item \textbf{0}: The object does not meaningfully reflect the background style, or the result is severely corrupted by excessive style leakage, making the object visually incoherent or unnaturally distorted.
\end{itemize}

\paragraph{Composition and Harmonization.}

\textbf{Question.}
Does the object inside the provided bounding box in the model output look naturally integrated and visually plausible? Focus primarily on the cropped output region, and use the full model output only as secondary global context. Consider whether the insertion looks natural rather than pasted, and whether local lighting, shadow consistency, texture transition, and overall harmonization appear convincing. Also consider physical plausibility, such as whether the object interacts naturally with its surroundings (e.g., ground or supporting surfaces), whether spatial relationships and depth ordering are coherent, whether the object is fully and coherently rendered without missing parts, and whether occlusion relationships with surrounding elements are handled realistically.

\textbf{Scoring Guide (0--3).}
\begin{itemize}
    \item \textbf{3}: The cropped output region looks very natural and well integrated. The insertion does not appear pasted, and local lighting, shadow consistency, texture transition, and overall harmonization are convincing.
    \item \textbf{2}: The region is mostly natural, but there are minor issues in local consistency or harmonization. Slight artificiality may be noticeable.
    \item \textbf{1}: The region looks noticeably unnatural. Pasted appearance or weak local harmonization make the insertion feel inconsistent.
    \item \textbf{0}: The region clearly fails. It strongly looks like an artificial copy-and-paste insertion, with severe artifacts or implausible local appearance.
\end{itemize}

\paragraph{Task Description.}
\begin{quote}
Evaluate how well the generated object is preserved, stylized, and harmonized inside the target region. Use the criterion-specific image priorities described above.
\end{quote}

\paragraph{Output Format.}
\begin{quote}
\texttt{Identity Preservation: [score, 0--3] -- [brief justification]} \\
\texttt{Style Transfer Consistency: [score, 0--3] -- [brief justification]} \\
\texttt{Composition and Harmonization: [score, 0--3] -- [brief justification]} \\
\texttt{Total Score: [sum of the three scores]}
\end{quote}

\subsection{VLM-based Filtering}
\label{sec:vlm_filtering}
\paragraph{System / Role.}
\begin{quote}
You are a strict visual evaluator.
\end{quote}

\paragraph{Setup.}
After candidate generation, each candidate is associated with two images:
\textbf{Image~A}, the original crop containing the target object together
with surrounding occluders, and \textbf{Image~B}, the segmented foreground
produced by SAM~3, where occluders such as hands are intentionally removed.
The segmented region can therefore be small or fragmented, but a small region
alone is not a reason for rejection.
We use \textsc{Qwen3-VL-8B-Instruct} as the filter with greedy decoding
(\texttt{do\_sample=False}) and \texttt{max\_new\_tokens}$=512$.
Both images are resized so that the longer side is at most $256$ px before
being passed to the model.

\paragraph{Provided Inputs.}
The model receives:
\begin{enumerate}
    \item \textbf{Image~A}: the original crop image containing the target object and surrounding occluders.
    \item \textbf{Image~B}: the segmented foreground image of the target object.
    \item \textbf{Target label}: the semantic category associated with the candidate.
\end{enumerate}

\paragraph{Task Description.}
The filtering objective is to determine whether Image~B provides a useful
partial observation of the target object under occlusion.
The model evaluates whether the remaining visible evidence supports plausible
restoration or completion in later stages.

\paragraph{Evaluation Rules.}
\begin{itemize}
    \item The segmentation intentionally excludes occluders, so Image~B may contain only a partial object region.
    \item A small segmented region alone should not trigger rejection.
    \item The evaluation should focus on semantic consistency, occlusion plausibility, and recoverability from the visible evidence.
    \item The model must return only a valid JSON object.
    \item Markdown formatting and code fences are explicitly forbidden.
    \item We apply up to two retries on JSON parse failure.
\end{itemize}

\paragraph{Criteria.}
Each candidate is evaluated using four criteria:

\textbf{label\_consistency.}
Does Image~B still depict the target semantic class?

\textbf{identity\_preservation.}
Is the same object instance recognisable across Image~A and Image~B?

\textbf{occlusion\_plausibility.}
Is the missing region physically consistent with a realistic occluder?

\textbf{recoverability.}
Can the full object be plausibly reconstructed from the visible evidence in Image~B?

\paragraph{Aggregation.}
We aggregate the four scores into a single keep score
$s \in [0,100]$:
\[
\mathrm{sem}
=
0.35 \cdot \texttt{label\_consistency}
+
0.65 \cdot \texttt{identity\_preservation},
\]
\[
s
=
\Big\lfloor
100 \cdot \Big(
0.60 \cdot \texttt{recoverability}
+
0.25 \cdot \texttt{occlusion\_plausibility}
+
0.15 \cdot \mathrm{sem}
\Big)
\Big\rceil .
\]

Recoverability receives the largest weight because the downstream task
requires plausible amodal completion from partial observations.

\paragraph{Binary Decision.}
We convert the continuous score into a binary keep/reject label:
\[
\mathrm{keep}(s,\texttt{recoverability})
=
\mathds{1}
\left[
s \geq 70
\;\wedge\;
\texttt{recoverability} \geq 0.35
\right].
\]

Candidates with low recoverability are rejected regardless of the final
aggregated score because their visible regions do not provide sufficient
evidence for plausible completion.

\paragraph{Output Format.}
The model returns a single JSON object containing:
\begin{itemize}
    \item \texttt{label\_consistency}
    \item \texttt{identity\_preservation}
    \item \texttt{occlusion\_plausibility}
    \item \texttt{recoverability}
    \item \texttt{final\_score}
    \item \texttt{short\_reason}
\end{itemize}

For each candidate, we store the four sub-scores, the aggregated keep score,
and the final binary decision in the per-scene \texttt{result.json}.

\section{Limitations and Broader Impacts}
\label{app:statements}

\subsection{Limitations}
Although Chameleon generalizes across diverse source and target domains, unusual background appearances, such as strong defocus blur, can still make seamless integration challenging. In such cases, the foreground must adapt to the background appearance while also reflecting the degradation itself, which is not explicitly modeled by our framework. As a result, the foreground can preserve its identity yet remain insufficiently matched to the degraded background. This challenge is shared by existing cross-domain compositing methods, which often fail to preserve both foreground identity and background-consistent appearance under such conditions.

\begin{figure}[H]
    \centering
    \includegraphics[width=\linewidth]{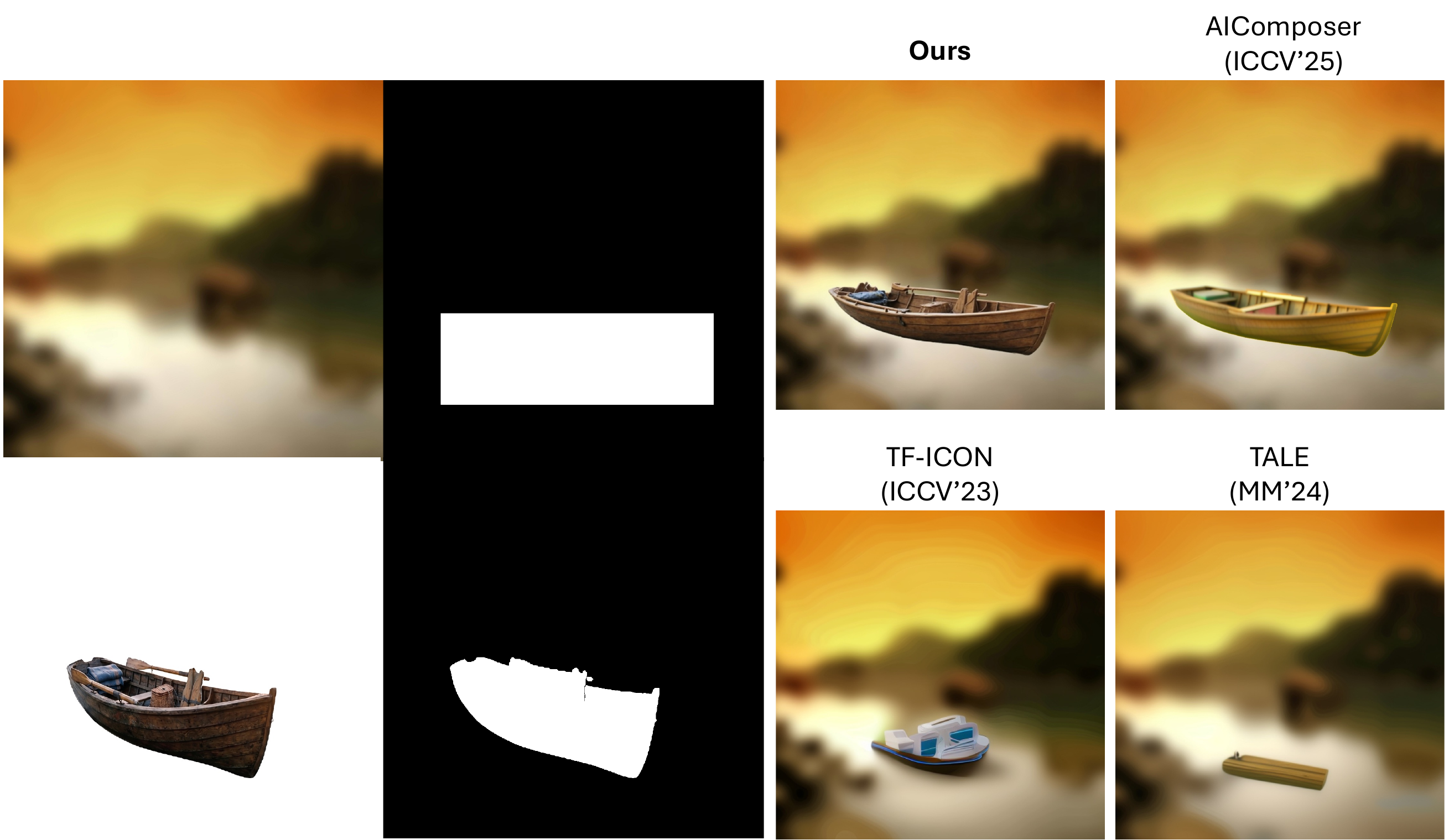}
    \vspace{-15pt}
    \caption{\textbf{Limitation on unseen degraded appearances.}
    When the background contains strong defocus blur, Chameleon preserves foreground identity but may not fully reproduce the background degradation, while existing cross-domain compositing methods also struggle to maintain both identity and background-consistent appearance.}
    \label{fig:limitation}
\end{figure}

\subsection{Broader Impacts}
Our work advances cross-domain image compositing by enabling more seamless and stylistically consistent integration between foreground objects and heterogeneous background domains. This capability may benefit creative applications such as digital art, visual content creation, and stylized media editing. However, the proposed framework may also be misused to generate misleading or deceptive visual content. In particular, realistic cross-domain compositing could be applied to create fabricated images that appear visually plausible, potentially contributing to misinformation or unauthorized image manipulation. In addition, generated results may raise concerns regarding copyright and artistic style imitation when specific visual styles are closely replicated.

\end{document}